\documentclass{article}
\usepackage[nonatbib, final]{neurips_2018}
\usepackage[numbers]{natbib}
\usepackage[utf8]{inputenc} % allow utf-8 input
\usepackage[T1]{fontenc}    % use 8-bit T1 fonts
\usepackage[hidelinks]{hyperref}       % hyperlinks
\usepackage{url}            % simple URL typesetting
\usepackage{booktabs}       % professional-quality tables
\usepackage{amsfonts}       % blackboard math symbols
\usepackage{nicefrac}       % compact symbols for 1/2, etc.
\usepackage{microtype}      % microtypography
\usepackage{subcaption}
\usepackage{algorithm}
\usepackage{algorithmic}
\usepackage[utf8]{inputenc}
\usepackage{amssymb,relsize}
\usepackage[intlimits]{amsmath}
\usepackage{array}
\usepackage{multirow}
\usepackage{mathabx}
\usepackage{booktabs}
\usepackage{appendix}
\usepackage{placeins}
\usepackage{amsthm}
\usepackage{url}
\usepackage{enumitem}
\usepackage{subfiles}
\usepackage{caption}
\usepackage{graphicx,epstopdf} 
\usepackage{tikz}
\usetikzlibrary{matrix,chains,positioning,decorations.pathreplacing,arrows}
\usepackage{cleveref}% Load LAST

\newtheorem{theorem}{Theorem}
\newtheorem{proposition}{Proposition}

\newcommand{\mat}[1]{\ensuremath \boldsymbol{\mathbf{#1}}} % this will give upright characters for arabic letters
\newcommand{\order}[1]{$\mathcal{O}(#1)$}
\newcommand{\inR}[1]{\ensuremath \in \mathbb{R}^{#1}}
\newcommand{\W}{\mat{W}}

\newcommand{\mPhi}{\ensuremath \mat{\Phi}}

\newenvironment{varalgorithm}[1] 
  {\algorithm\renewcommand{\thealgorithm}{#1}}
  {\endalgorithm}

\newcommand{\logl}{\log \mat{\ell}}
\newcommand{\logp}{\log \mat{p}}
\newcommand{\logq}{\log \mat{q}}

\newcommand{\lesslines}{\looseness=-1} % Try one fewer line for the following paragraph

\newcommand{\conditionalBib}{\dobib}

\setlist[itemize]{leftmargin=*}

\newcommand{\mytitle}{Discretely Relaxing Continuous Variables\\ for tractable Variational Inference}
\title{\mytitle}

\author{  
	Trefor W. Evans \\
	University of Toronto\\
	\texttt{trefor.evans@mail.utoronto.ca}
	\And
	Prasanth B. Nair \\
	University of Toronto \\
	\texttt{pbn@utias.utoronto.ca}
}

\begin{document}
\renewcommand{\conditionalBib}{} % undefine this when compiling the main document
\maketitle
\begin{abstract}
We explore a new research direction in Bayesian variational inference with discrete latent variable priors
where we exploit Kronecker matrix algebra for efficient and exact computations of the evidence lower bound~(ELBO).
The proposed "DIRECT" approach has several advantages over its predecessors; 
(i)~it can exactly compute ELBO gradients (i.e. unbiased, zero-variance gradient estimates), eliminating the need for high-variance stochastic gradient estimators and enabling the use of quasi-Newton optimization methods; 
(ii)~its training complexity is \emph{independent} of the number of training points, permitting inference on large datasets; and
(iii)~its posterior samples consist of sparse and low-precision quantized integers which permit fast inference on hardware limited devices.
In addition, our DIRECT models can exactly compute statistical moments of the parameterized predictive posterior without relying on Monte Carlo sampling.
The DIRECT approach is not practical for all likelihoods, however, we identify a popular model structure which is practical, and 
demonstrate accurate inference using latent variables discretized as extremely low-precision 4-bit quantized integers.
While the ELBO computations considered in the numerical studies require over $10^{2352}$ log-likelihood evaluations, 
we train on datasets with over two-million points in just seconds.
%We demonstrate efficient learning and inference on several popular model structures and discuss how the techniques can
% https://www.wolframalpha.com/input/?i=15%5E2000
\end{abstract}

\section{Introduction}
Hardware restrictions posed by mobile devices make Bayesian inference particularly ill-suited for on-board machine learning. 
This is unfortunate since the safety afforded by Bayesian statistics is extremely valuable in many prominent mobile applications.
For example, the cost of erroneous decisions are very high in autonomous driving or mobile robotic control.
The robustness and uncertainty quantification provided by Bayesian inference is therefore extremely valuable for these applications provided inference can be performed on-board in real-time~\cite{thrun_probabilistic_robotics,ghahramani_adversarial}.

Outside of mobile applications, resource efficiency is still an important concern.
For example, deployed models making billions of predictions per day can incur substantial energy costs, making energy efficiency an important consideration in modern machine learning architectures~\cite{welling_bayes_compression}.

We approach the problem of efficient Bayesian inference by considering discrete latent variable models such that posterior samples of the variables will be quantized and sparse, leading to efficient inference computations with respect to energy, memory and computational requirements.
Training a model with a discrete prior is typically very slow and expensive, requiring the use of high variance Monte Carlo gradient estimators to learn the variational distribution.
The main contribution of this work is the development of a method to rapidly learn the variational distribution for such a model without the use of any stochastic estimators; the objective function will be computed exactly at each iteration.
To our knowledge, such an approach has not been taken for variational inference of large-scale probabilistic models.

In this paper, we compare our work not only to competing stochastic variational inference (SVI) methods for discrete latent variables, but also to the more general SVI methods for continuous latent variables.
We make this comparison with continuous variables by discretely relaxing continuous priors using a discrete prior with a finite support set that contains much of the structure and information as its continuous analogue.
Using this discretized prior we show that we can make use of Kronecker matrix algebra for efficient and exact ELBO computations.
We will call our technique DIRECT~(DIscrete RElaxation of ConTinous variables).
We summarize our main contributions below:
\begin{itemize}
	\item We efficiently and exactly compute the ELBO using a discrete prior even when this computation requires more likelihood evaluations than the number of atoms in the known universe. 
	This achieves unbiased, zero-variance gradients which we show outperforms competing Monte Carlo sampling alternatives that give high-variance gradient estimates while learning.
	\item Complexity of our ELBO computations are \emph{independent} of the quantity of training data using the DIRECT method, making the proposed approach amenable to big data applications.
	\item At inference time, we can exactly compute the statistical moments of the parameterized predictive posterior distribution, unlike competing techniques which rely on Monte Carlo sampling.
	\item Using a discrete prior, our models admit sparse posterior samples that can be represented as quantized integer values to enable efficient inference, particularly on hardware limited devices.
	\item We present the DIRECT approach for generalized linear models and deep Bayesian neural networks for regression, and discuss approximations that allow extensions to many other models.
	%Our focus is on mean-field variational inference, however, we demonstrate how an unfactorized variational distribution can be used by introducing a manageable level of stochasticity into the gradients.
	%\item We discuss extensions of the DIRECT method that enable efficient ELBO lower bound computations, allowing extensions of the proposed techniques to scenarios where a compact representation cannot be found.
	\item Our empirical studies demonstrate superior performance relative to competing SVI methods on problems with as many as 2~million training points.
\end{itemize}

The paper will proceed as follows;
\cref{sec:vi} contains a background on variational inference and poses the learning problem to be addressed while
\cref{sec:inference} outlines the central ideas of the DIRECT method, demonstrating the approach on several popular probabilistic models.
\Cref{sec:limitations} discusses limitations of the proposed approach and outlines some work-arounds, for instance, we discuss how to go beyond mean-field variational inference.
We empirically demonstrate our approaches in \cref{sec:studies}, and conclude in \cref{sec:conclusions}.
Our full code is available at \url{https://github.com/treforevans/direct}.

\section{Variational Inference Background}
\label{sec:vi}
\label{sec:background}
We begin with a review of variational inference, a method for approximating probability densities in Bayesian statistics~\cite{jordan_vi_root, wainwright_vi_root, hoffman_svi, ranganath_blackbox_vi, blei_blackbox_vi, blei_vi_review}.
We introduce a regression problem for motivation;
given $\mat{X} \inR{n\times d},\ \mat{y}\inR{n}$, a $d$-dimensional dataset of size $n$, we wish to evaluate $y_*$ at an untried point $\mat{x}_*$ by constructing a statistical model that depends on the $b$ latent variables in the vector $\mat{w} \inR{b}$.
After specifying a prior over the latent variables, $\Pr(\mat{w})$, and selecting a probabilistic model structure that admits the likelihood $\Pr(\mat{y}|\textbf{w})$, we may proceed with Bayesian inference to determine the posterior $\Pr(\mat{w}|\mat{y})$ which generally requires analytically intractable computations.

Variational inference turns the task of computing a posterior into an optimization problem.
By introducing a family of probability distributions $q_{\mat{\theta}}(\mat{w})$ parameterized by $\mat{\theta}$, we minimize the Kullback-Leibler divergence to the exact posterior~\cite{blei_vi_review}.
This equates to maximization of the evidence lower bound (ELBO) which we can write as follows for a continuous or discrete prior, respectively
\vspace{1.5mm}
\begin{minipage}[t]{0.25\textwidth}
\centering
Prior
\includegraphics[width=\textwidth,height=1cm]{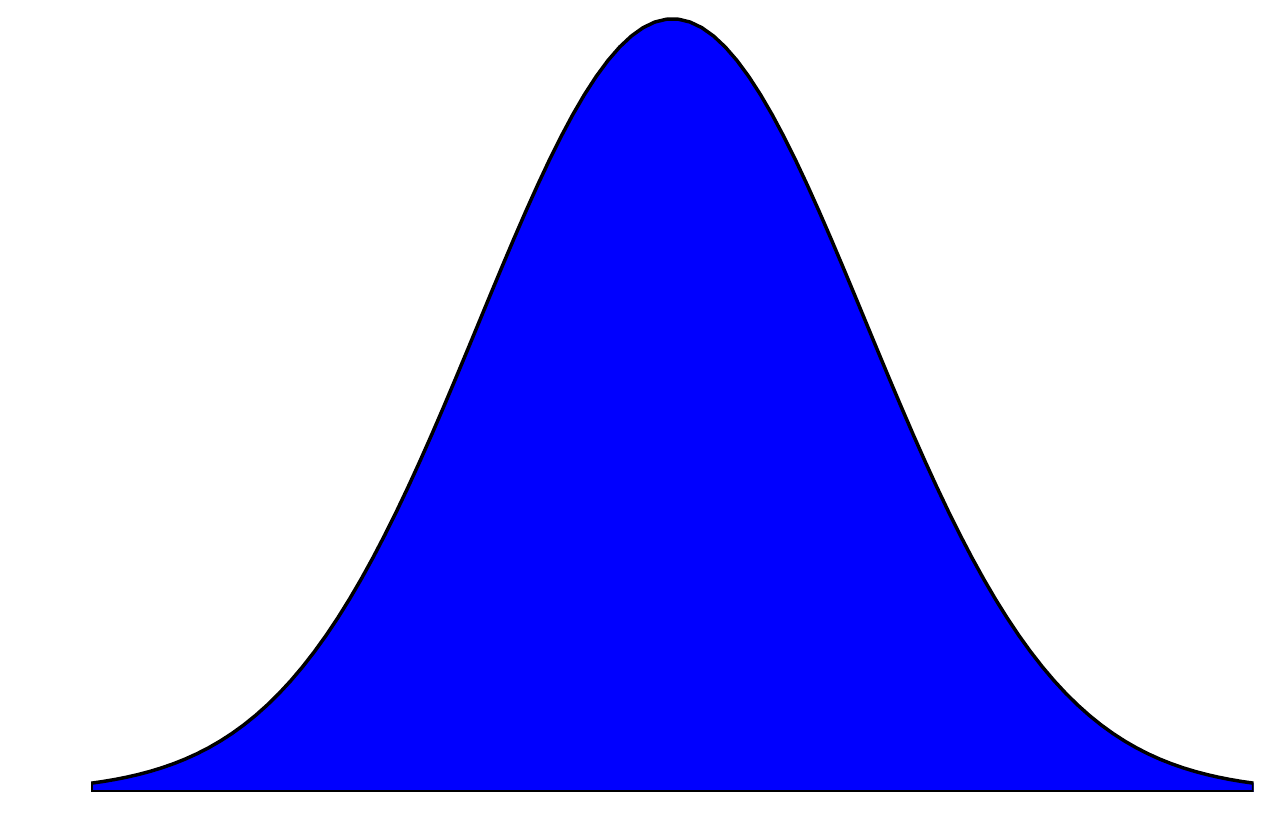}\\
\includegraphics[width=\textwidth,height=1cm]{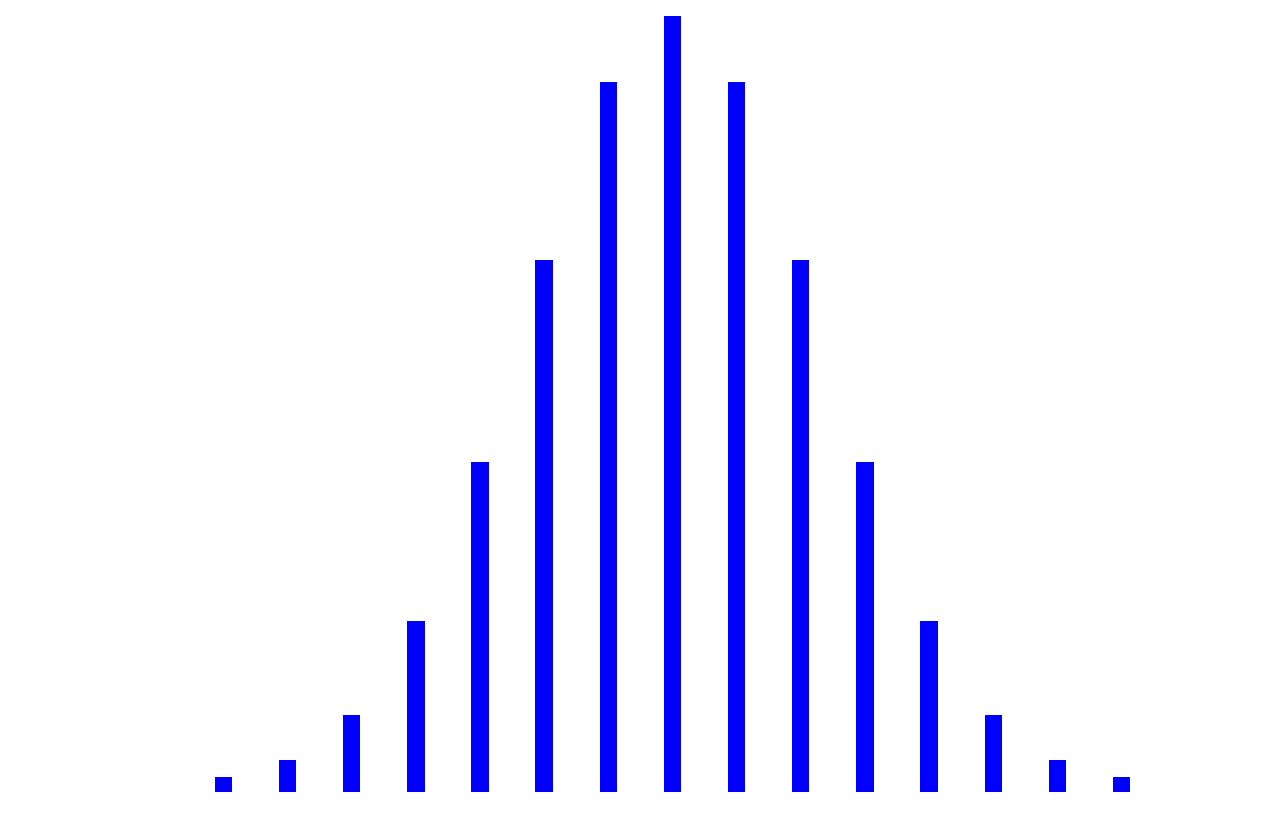}
\end{minipage}%
\begin{minipage}[t]{0.75\textwidth}
\centering
ELBO\\[-6mm]
\begin{align}
\label{eqn:ELBO_continuous}
\textsf{ELBO}(\mat{\theta})
&= \int q_{\mat{\theta}}(\textbf{w}) \Big(\log \Pr(\mat{y}|\textbf{w}) + \log\Pr(\textbf{w}) - \log q_{\mat{\theta}}(\mat{w})\Big) d\textbf{w},
\\[3mm]
\label{eqn:ELBO}
\textsf{ELBO}(\mat{\theta})
&= \mat{q}^T\Big(\logl + \logp - \logq\Big),
\end{align}
\end{minipage}
\vspace{1.5mm}
where 
$\logl = \{\log \Pr(\mat{y}|\textbf{w}_i)\}_{i=1}^m$,
$\logp = \{\log \Pr(\textbf{w}_i)\}_{i=1}^m$,
$\mat{q} = \{q_{\mat{\theta}}(\mat{w}_i)\}_{i=1}^m$, and 
$\{\mat{w}_i\}_{i=1}^m = \mat{W} \inR{b \times m}$ is the entire support set of the discrete prior.

It is immediately evident that computing the ELBO is challenging when $b$ is large, since 
in the continuous case \cref{eqn:ELBO_continuous} is a $b$-dimensional integral, and
in the discrete case the size of the sum in \cref{eqn:ELBO} generally increases exponentially with respect to $b$.
Typically, the ELBO is not explicitly computed and instead, a Monte Carlo estimate of the gradient of the ELBO with respect to the variational parameters $\mat{\theta}$ is found, allowing stochastic gradient descent to be performed.
We will outline some existing techniques to estimate ELBO gradients with respect to the variational parameters, $\mat{\theta}$.

For continuous priors, the reparameterization trick \cite{kingma_reparam} can be used to perform variational inference.
The technique uses Monte Carlo estimates of the gradient of the evidence lower bound (ELBO) which is maximized during the training procedure.
While this approach has been employed successfully for many large-scale models, we find that discretely relaxing continuous latent variable priors can improve training and inference performance when using our proposed DIRECT technique which computes the ELBO (and its gradients) exactly.

When the latent variable priors are discrete, reparameterization cannot be applied, however, the REINFORCE~\cite{williams_reinforce} estimator may be used to provide an unbiased estimate of the ELBO during training~(alternatively called the score function estimator~\cite{fu_score_function}, or likelihood ratio estimator~\cite{glynn_likelihood_ratio}).
Empirically, the REINFORCE gradient estimator is found to give a high-variance when compared with reparameterization, leading to a slow learning process.
Unsurprisingly, we find that our proposed DIRECT technique trains significantly faster than a model trained using a REINFORCE estimator.

\lesslines
Recent work in variational inference with discrete latent variables has largely focused on continuous relaxations of discrete variables such that reparameterization can be applied to reduce gradient variance compared to REINFORCE.
One example is CONCRETE~\cite{jang_gumbel_softmax, maddison_concrete} and its extensions~\cite{tucker_rebar, grathwohl_relax}.
We consider an opposing direction by identifying how the ELBO~(\cref{eqn:ELBO}) can be computed exactly for a class of discretely relaxed probabilistic models such that the discrete latent variable model can be trained more easily then its continuous counterpart.
We outline this approach in the following section.
%We call this technique DIRECT which we outline in the following section.
%While the number of terms in this sum increases exponentially in $b$, we will see that through the use of Kronecker matrix algebra we can reduce this complexity dramatically for many models.

\section{DIRECT: Efficient ELBO Computations with Kronecker Matrix Algebra}
\label{sec:inference}
\label{sec:direct}
\lesslines
We outline the central ideas of the DIRECT method and illustrate its application on several probabilistic models.
The DIRECT method allows us to efficiently and exactly compute the ELBO which has several advantages over existing SVI techniques for discrete latent variable models such as, zero-variance gradient estimates, the ability to use a super-linearly convergent quasi-Newton optimizer (since our objective is deterministic), and the per-iteration complexity is independent of training set size.
We will also discuss advantages at inference time such as the ability to exactly compute predictive posterior statistical moments, and to exploit sparse and low-precision posterior samples.

To begin, we consider a discrete prior over our latent variables whose support set $\mat{W}$ forms a Cartesian tensor product grid as most discrete priors do (e.g. any prior that factorizes between variables) so that we can write
\begin{equation} \label{eqn:w_grid}
\mat{W}
= \left(\begin{array}{ccccccc}
\widebar{\mat{w}}_1^T &\otimes& \mat{1}_{\widebar{{m}}}^T &\otimes& \cdots &\otimes& \mat{1}_{\widebar{{m}}}^T\\
\mat{1}_{\widebar{{m}}}^T &\otimes& \widebar{\mat{w}}_2^T &\otimes& \cdots &\otimes& \mat{1}_{\widebar{{m}}}^T\\
\vdots && \vdots && \ddots && \vdots\\
\mat{1}_{\widebar{{m}}}^T &\otimes& \mat{1}_{\widebar{{m}}}^T &\otimes& \cdots &\otimes& \widebar{\mat{w}}_b^T
\end{array} \right),
\end{equation}
where 
$\mat{1}_{\widebar{m}} \inR{\widebar{m}}$ denotes a vector of ones,
$\widebar{\mat{w}}_i \inR{\widebar{m}}$ contains the $\widebar{m}$ discrete values that the $i$th latent variable $w_i$ can take%
\footnote{The discrete values that the $i$th latent variable can take, $\widebar{\mat{w}}_i$, may be chosen a priori or learned during ELBO maximization (may be helpful for coarse discretizations).
For the sake of simplicity, we focus on the former.}, $m = \widebar{m}^b$, and $\otimes$ denotes the Kronecker product~\cite{van_loan_kron}.
Since the number of columns of $\mat{W} \inR{b \times \widebar{m}^b}$ increases exponentially with respect to $b$, it is evident that computing the ELBO in \cref{eqn:ELBO} is typically intractable when $b$ is large.
For instance, forming and storing the matrices involved naively require exponential time and memory.
We can alleviate this concern if $\mat{q}$, $\logp$, $\logl$, and $\logq$ can be written as a sum of Kronecker product vectors~(i.e.
$\sum_{i} \bigotimes_{j=1}^b \mat{f}_{j}^{(i)}$, where $\mat{f}_j^{(i)} \inR{\widebar{m}}$).
If we find this structure, then we never need to explicitly compute or store a vector of length $m$.
This is because \cref{eqn:ELBO} would simply require multiple inner products between Kronecker product vectors which the following result demonstrates can be computed extremely efficiently.
\begin{proposition}
The inner product between two Kronecker product vectors 
$\mat{k} = \otimes_{i=1}^b \mat{k}^{(i)}$, and
$\mat{a} = \otimes_{i=1}^b \mat{a}^{(i)}$
can be computed as follows~\cite{van_loan_kron},
\begin{align}
\mat{a}^T \mat{k} = \prod_{i=1}^b \mat{a}^{(i)\,T} \mat{k}^{(i)},
\end{align}
where
$\mat{a}^{(i)} \inR{\widebar{m}}$, $\mat{a} \inR{\widebar{m}^b}$, 
$\mat{k}^{(i)} \inR{\widebar{m}}$, and $\mat{k} \inR{\widebar{m}^b}$.
\end{proposition}
This result enables substantial savings in the computation of the ELBO since each inner product computation is reduced from the naive \emph{exponential} \order{\widebar{m}^b} cost to a \emph{linear} \order{b\widebar{m}} cost.

We now discuss how the Kronecker product structure of the variables in \cref{eqn:ELBO} can be achieved.
Firstly, if the prior is chosen to factorize between latent variables, as it often is, (i.e. 
$\Pr(\mat{w}) = \prod_{i=1}^b \Pr(w_i)$) then
$\mat{p} = \otimes_{i=1}^b \mat{p}_i$ admits a Kronecker product structure where
$\mat{p}_i = \{\Pr(w_i{=}\widebar{{w}}_{ij})\}_{j=1}^{\widebar{m}} \in (0,1)^{\widebar{m}}$.
The following result demonstrates how this structure for $\mat{p}$ enables $\logp$ to be written as a sum of $b$ Kronecker product vectors.
\begin{proposition} \label{prop:kron_sum}
The element-wise logarithm of the Kronecker product vector $\mat{k} = \otimes_{i=1}^b \mat{k}^{(i)}$ can be written as a sum of $b$ Kronecker product vectors as follows,
\begin{align}
\log \mat{k} = \bigoplus_{i=1}^b \log \mat{k}^{(i)},
\end{align}
where $\mat{k}^{(i)} \inR{\widebar{m}}$, $\mat{k} \inR{\widebar{m}^b}$ contain positive values, and
$\oplus$ is a generalization of the Kronecker sum~\cite{kron_sum} for vectors which we define as follows
\begin{align}
\bigoplus_{i=1}^b \log \mat{k}^{(i)} = \sum_{i=1}^b 
\bigg( \bigotimes_{j=1}^{i-1} \mat{1}_{\widebar{m}} \bigg)
\otimes
\log \mat{k}^{(i)}
\otimes
\bigg( \bigotimes_{j=i+1}^{b} \mat{1}_{\widebar{m}} \bigg).
\end{align}
\end{proposition}
The proof is trivial.
We will first consider a mean-field variational distribution that factorizes over latent variables such that both
$\mat{q} = \otimes_{i=1}^b \mat{q}_i$ and $\logq = \oplus_{i=1}^b \logq_i$ can be written as a sum of Kronecker product vectors, where
$\mat{q}_j = \{\Pr(w_j{=}\widebar{w}_{ji})\}_{i=1}^{\widebar{m}} \in (0,1)^{\widebar{m}}$ are used as the variational parameters, $\mat{\theta}$, with the use of the softmax function.
For the mean-field case we can rewrite \cref{eqn:ELBO} as
\begin{align} \label{eqn:ELBO2}
\textsf{ELBO}(\mat{\theta})
= \mat{q}^T\logl 
+ \sum_{i=1}^b \mat{q}_i^T \logp_i 
- \sum_{i=1}^b \mat{q}_i^T \logq_i ,
\end{align}
where we use the fact that $\mat{q}_i$ defines a valid probability distribution for the $i$th latent variable such that
$\mat{q}_i^T \mat{1}_{\widebar{m}} = 1$.
We extend results to unfactorized prior and variational distributions later in \cref{sec:correlated_q_lb}.

The structure of $\logl$ depends on the probabilistic model used;
in the worst case, $\logl$ can always be represented as a sum of $m$ Kronecker product vectors. 
However, many models admit a far more compact structure where dramatic savings can be realized as we demonstrate in the following sections.

\subsection{Generalized Linear Regression}
\label{sec:glm}
We first focus on the popular class of Bayesian generalized linear models (GLMs) for regression.
While the Bayesian integrals that arise in GLMs can be easily computed in the case of conjugate priors, for general priors inference is challenging.

This highly general model architecture has been applied in a vast array of application areas.
Recently, \citet{wilson_deep} used a scalable Bayesian generalized linear model with Gaussian priors on the output layer of deep neural network with notable empirical success.
They also considered the ability to train the neural network simultaneously with the approximate Gaussian process which we also have the ability to do if a practitioner were to require such an architecture.
\begin{comment}
Further, while the algorithm employed by \citet{wilson_deep} was restricted to very low dimensionality problems with fewer than 5 input dimensions~\cite{wilson_product_kernel_interp}, the proposed technique is not restricted by this.
Also, we will see that storage and computation requirements for our technique are \emph{independent} of the number of training points as opposed to the linear scaling of the method of \citet{wilson_deep}, we are not restricted to conjugate priors, and we can exploit the computational advantages of sparse and quantized posterior samples, as we will discuss.
\end{comment}

\lesslines
Consider the generalized linear regression model
$\mat{y} = \mat{\Phi} \mat{w} + \mat{\epsilon}$, where
$\mat{\epsilon} \sim \mathcal{N}(\mat{0}, \sigma^2 \mat{I})$, and
$\mPhi = \{\phi_j(\mat{x}_i)\}_{i,j} \inR{n \times b}$ contains the evaluations of the basis functions on the training data.
The following result demonstrates how the ELBO can be exactly and efficiently computed, assuming the factorized prior and variational distributions over $\mat{w}$ discussed earlier.
Note that we also consider a prior over $\sigma^2$.

\begin{theorem} \label{thm:glm}
The ELBO can be exactly computed for a discretely relaxed regression GLM as follows
\begin{multline}
\textsf{ELBO}(\mat{\theta})
=-\frac{n}{2}\mat{q}_{\sigma}^T \log\mat{\sigma}^2
-\frac{1}{2}\big(\mat{q}_{\sigma}^T \mat{\sigma}^{-2}\big)
\Big( 
\mat{y}^T\mat{y}
- 2 \mat{s}^T \big(\mat{\Phi}^T \mat{y}\big)
+ \mat{s}^T\mat{\Phi}^T\mat{\Phi}\mat{s} - \textsf{diag}(\mat{\Phi}^T\mat{\Phi})^T\mat{s}^2 + \\ \sum_{j=1}^{b}  \mat{q}_j^T \mat{h}_j
\Big)
+ \sum_{i=1}^b \big(\mat{q}_i^T \logp_i 
- \mat{q}_i^T \logq_i\big)
+ \mat{q}_{\sigma}^T \logp_{\sigma}
- \mat{q}_{\sigma}^T \logq_{\sigma},
\end{multline}
where
$\mat{q}_{\sigma}, \mat{p}_{\sigma} \inR{\widebar{m}}$ are factorized variational and prior distributions over the Gaussian noise variance $\sigma^2$ for which we consider the discrete positive values $\mat{\sigma}^2 \inR{\widebar{m}}$, respectively.
Also, we use the shorthand notation
$\mat{H} = \{\widebar{\mat{w}}_j^2 \sum_{i=1}^n \phi_{ij}^2\}_{j=1}^b \inR{\widebar{m} \times b}$, and
$\mat{s} = \{\mat{q}_j^T\widebar{\mat{w}}_j\}_{j=1}^b \inR{b}$.
\end{theorem}
A proof is provided in \cref{sec:glm_proof} of the supplementary material.
We can pre-compute the terms $\mat{y}^T\mat{y}$, $\mat{\Phi}^T \mat{y}$, $\mat{H}$, and 
$\mat{\Phi}^T\mat{\Phi}$ before training begins (since these do not depend on the variational parameters)
such that the final complexity of the proposed DIRECT method outlined in \Cref{thm:glm} is only \order{b\widebar{m} + b^2}.
This complexity is \emph{independent} of the number of training points, making the proposed technique ideal for massive datasets.
Also, each of the pre-computed terms can easily be updated as more data is observed making the techniques amenable to online learning applications.

\paragraph{Predictive Posterior Computations}
Typically, the predictive posterior distribution is found by sampling the variational distribution at a large number of points and running the model forward for each sample.
To exactly compute the statistical moments, a model would have to be run forward at every point in the hypothesis space with is typically intractable, however, we can exploit Kronecker matrix algebra to efficiently compute these moments exactly.
For example, the exact predictive posterior mean for our generalized linear regression model is computed as follows
\begin{align}\label{eqn:post_mean}
\mathbb{E}(y_*) = \sum_{i=1}^m q(\mat{w}_i) \int y_*\Pr(y_*|\mat{w}_i) dy_*,
= \mat{\Phi}_* \mat{W} \mat{q} = \mat{\Phi}_* \mat{s},
\end{align}
where $\mat{s} = \{\mat{q}_j^T\widebar{\mat{w}}_j\}_{j=1}^b \inR{b}$, and 
$\mat{\Phi}_*\inR{1\times b}$ contains the basis functions evaluated at $x_*$.
This computation is highly efficient, requiring just \order{b} time per test point.
It can be shown that a similar scheme can be derived to exactly compute higher order statistical moments, such as the predictive posterior variance, for generalized linear regression models and other DIRECT models.

We have shown how to exactly compute statistical moments, and now we show how to exploit our discrete prior to compute predictive posterior samples extremely efficiently.
This sampling approach may be preferable to the exact computation of statistical moments on hardware limited devices where we need to perform inference with extreme memory, energy and computational efficiency.
The latent variable posterior samples $\widetilde{\mat{W}} \inR{b \times \text{num. samples}}$ will of course be represented as a low-precision quantized integer array because of the discrete support of the prior which enables extremely compact storage in memory.
Much work has been done elsewhere in the machine learning community to quantize variables for storage compression purposes since memory is a very restrictive constraint on mobile devices~\cite{chen_compressing_nn,gong_compressing_nn,han_compressing_nn,zhou_compression_nn}.
However, we can go beyond this to additionally reduce computational and energy demands for the evaluation of $\mat{\Phi}_*\widetilde{\mat{W}}$.
One approach is to constrain the elements of $\widebar{\mat{w}}$ to be 0 or a power of 2 so that multiplication operations simply become efficient bit-shift operations~\cite{bengio_binary_nn,li_ternary_nn,rastegari_xnor_nn}.
An even more efficient approach is to employ basis functions with discrete outputs so that $\mat{\Phi}_*$ can also be represented as a low-precision quantized integer array.
For example, a rounding operation could be applied to continuous basis functions.
Provided that the quantization schemes are an affine mapping of integers to real numbers (i.e. the quantized values are evenly spaced),
then inference can be conducted using extremely efficient integer arithmetic~\cite{jacob_quantization_nn}. \label{sec:int_arithmetic}
Either of these approaches enable extremely efficient on-device inference.

\subsection{Deep Neural Networks for Regression}
We consider the hierarchical model structure of a Bayesian deep neural network for regression.
Considering a DIRECT approach for this architecture is not conceptually challenging so long as an appropriate neuron activation function is selected.
We would like a non-linear activation that maintains a compact representation of the log-likelihood evaluated at every point in the hypothesis space, i.e. we would like $\logl$ to be represented as a sum of as few Kronecker product vectors as possible.
Using a power function for the activation can maintain a compact representation; the natural choice being a quadratic activation function (i.e. output $x^2$ for input $x$).

\lesslines
It can be shown that the ELBO can be exactly computed in \order{\ell \widebar{m}(b/\ell)^{4\ell}} for a deep Bayesian neural network with $\ell$ layers, where
we assume a quadratic activation function and 
an equal distribution of discrete latent variables between network layers.
This complexity evidently enables scalable Bayesian inference for models of moderate depth, and like we found for the regression GLM model of \cref{sec:glm}, computational complexity is \emph{independent} of the quantity of training data, making this approach ideal for large datasets. 
We outline this model and the computation of its ELBO in \cref{sec:bnn}.

\begin{comment} % high level overview
computing the using a quadratic activation function 
In \cref{sec:glm}, we saw how to compute $\mat{z}_i^2$ as a sum of $(b+b^2)/2$ Kronecker product vectors.
In the Bayesian neural network case, a similar computation can be performed for each node in the network, however, the output of each activation function will be a sum of $(\widetilde{b}+\widetilde{b}^2)/2$ Kronecker product vectors given a sum of $\widetilde{b}$ Kronecker product vectors as the input.
As a result, the ELBO computation time complexity for a Bayesian neural network model with $\ell$ layers will be \order{b\widebar{m} + (b/\ell)^{2\ell}}, assuming the latent variables are distributed equally between layers.
\end{comment}

\section{Limitations \& Extensions}
\label{sec:limitations}
\label{sec:extensions}
In generality, when the support of the prior is on a Cartesian grid, any prior, likelihood, or variational distribution (or log-distribution) can be expressed using the proposed Kronecker matrix representation, however, this representation will not always be compact enough to be practical.
We can see this by viewing these probability distributions over the hypothesis space as high-dimensional tensors.
In \cref{sec:direct}, we exploited some popular models
whose variational probability tensors, and
whose prior, likelihood and variational log-probability tensors all admit a low-rank structure, however,
other models may not admit this structure, in which case their representation will not be so compact.
In the interest of generalizing the technique, we outline a likelihood, a prior, and a variational distribution that does not admit a compact representation of the ELBO and discuss several ways the DIRECT method can still be used to efficiently compute, or lower bound the ELBO.
We hope that these extensions inspire future research directions in approximate Bayesian inference.

\paragraph{Generalized Linear Logistic Regression}
\lesslines
Logistic regression models do not easily admit a compact representation for exact ELBO computations, however, we will demonstrate that we can efficiently compute a lower-bound of the ELBO by leveraging developed algebraic techniques. 
To demonstrate, we will consider a generalized linear logistic regression model which is commonly employed for classification problems.
Such a model could easily be extended to a deep architecture following \citet{ghahramani_adversarial}, if desired.
All terms in the ELBO in \cref{eqn:ELBO2} can be computed exactly for this model except the term involving the log-likelihood, for which the following result demonstrates an efficient computation of the lower bound.
\begin{theorem} \label{thm:logistic}
For a generalized linear logistic regression model with
classification training labels $\mat{y} \in \{0,1\}^n$,
the class-conditional probability $\Pr(y_i{=}0|\mat{w}) = (1+\exp(-\mPhi[i,:]\mat{w}))^{-1}$,
and with the assumption that training examples are sampled independently,
%a factorized prior and variational distribution,
the following inequality holds
\begin{align}
\mat{q}^T\logl \geq
-\mat{s}^T \big(\mat{\Phi}^T \mat{y}\big) 
-\sum_{i=1}^n \left\{ \begin{array}{ll}
\prod_{j=1}^b \mat{q}_j^T\exp(-\phi_{ij} \widebar{\mat{w}}_j)&\text{if}\ y_i = 0\\
\prod_{j=1}^b \mat{q}_j^T\exp(\phi_{ij} \widebar{\mat{w}}_j)-\sum_{j=1}^b\mat{q}_i^T\phi_{ij} \widebar{\mat{w}}_j&\text{if}\ y_i = 1
\end{array}\right.
\end{align}
\end{theorem}
We prove this result in \cref{sec:logistic_proof} of the supplement.
This computation can be performed in \order{\widebar{m}bn} time, where dependence on $n$ is evident unlike in the case of the exact computations described in \cref{sec:direct}.
As a result, stochastic optimization techniques should be considered.
Using this lower bound, the log-likelihood is accurately approximated for hypotheses that correctly classify the training data, however, hypotheses that confidently misclassify training labels may be over-penalized.
In \cref{sec:logistic_remarks} we further discuss the accuracy of this approximation and discuss a stable implementation.

\paragraph{Unfactorized Variational Distributions}
\label{sec:correlated_q_lb} 
We now consider going beyond a mean-field variational distribution to account for correlations between latent variables.
Considering a finite mixture of factorized categorical distributions as is used in latent structure analysis~\cite{lazarsfeld_lsa,goodman_lsa}, % same ref made in https://www.ncbi.nlm.nih.gov/pmc/articles/PMC3630378/ just before eqn 1
we can write
$\mat{q} = \sum_{i=1}^r \alpha_i \bigotimes_{j=1}^b \mat{q}_j^{(i)}$,
where
$\mat{\alpha} \in (0,1)^{r}$ is a vector of mixture probabilities for $r$ components, and
$\mat{q}_j^{(i)} = \{\Pr(w_j{=}\widebar{w}_{jk}|i)\}_{k=1}^{\widebar{m}} \in (0,1)^{\widebar{m}}$.
\begin{comment}
Observe that we are essentially representing the matrix of probabilities of a $b$-dimensional categorical distribution as a sum of Kronecker product matrices.
This structure is a well-known compact representation of matrices studied in tensor decomposition literature~\cite{refs}.
As a result, we arrive with a highly compressed and flexible variational parameterization.
\end{comment}

While $\mat{q}$ can evidently be expressed as a compact sum of Kronecker product vectors, $\logq$ is more challenging to compute than in the mean-field case, however, the following result demonstrates how we can lower-bound the term involving $\logq$ in the ELBO (\cref{eqn:ELBO2}).
\begin{theorem} \label{thm:mixture_lb}
The following inequality holds when we consider a finite mixture of factorized categorical distributions for $q_{\mat{\theta}}(\mat{w})$,
\begin{align} \nonumber
\hspace{-3mm}
{-}\mat{q}^T\logq \geq 
\hspace{-1mm}
\underset{\{\mat{a}_i \in (0,1)^{\widebar{m}}\}_{i=1}^b}{\text{max}}
\hspace{-1mm}
1-\sum_{j=1}^r \alpha_j\bigg(\sum_{i=1}^b 
\mat{q}^{(j)\,T}_i \log \mat{a}_i 
+ \alpha_j\prod_{i=1}^b \mat{q}_i^{(j)\,T} \frac{\mat{q}_i^{(j)}}{\mat{a}_i}
+ 2\hspace{-2mm}\sum_{k{=}j{+}1}^r \hspace{-2mm}\alpha_k\prod_{i=1}^b \mat{q}_i^{(j)\,T} \frac{\mat{q}_i^{(k)}}{\mat{a}_i}
\bigg),
\end{align}
where $\mat{a} = \otimes_{i=1}^b \mat{a}_i$, $\mat{a}_i \in (0,1)^{\widebar{m}}$ is the center of the Taylor series approximation of $\logq$.
\end{theorem}
We prove this result in \cref{sec:mixture_proof} and discuss a stable implementation.
Note that if the mixture variational distribution $\mat{q}$ degenerates to a mean-field distribution equal to $\mat{a}$, then the ELBO will be computed exactly, and as $\mat{q}$ moves away from $\mat{a}$, the ELBO will be underestimated.
%The effect of the approximation therefore introduces a false affinity to mean-field solutions, however, we find in our studies that this does not prevent the discovery of meaningful posterior correlations.

\paragraph{Unfactorized Prior Distributions}
To consider an unfactorized prior, we assume a prior mixture distribution given by
$\mat{p}= \sum_{i=1}^r \alpha_i \bigotimes_{j=1}^b \mat{p}_j^{(i)}$.
When we use this mixture distribution for the prior, $\mat{p}$ can evidently be expressed as a compact sum of Kronecker product vectors but $\logp$ cannot. 
The following result demonstrates how we can still lower-bound the term involving $\logp$ in the ELBO (\cref{eqn:ELBO}).
For simplicity, we assume that the variational distribution factorizes, however, the result could easily be extended to the case of a mixture variational distribution.
\begin{theorem} \label{thm:mixture_prior}
The following inequality holds when we consider a finite mixture of factorized categorical distributions for $p_{\mat{\theta}}(\mat{w})$,
\begin{align} \nonumber
\mat{q}^T\logp 
\geq 
%\sum_{i=1}^r \alpha_i \mat{q}^T\logp^{(i)} =
\sum_{i=1}^r \alpha_i \sum_{j=1}^b \mat{q}_j^T\logp_j^{(i)}
\end{align}
\end{theorem}
The proof is trivial by Jensen's inequality.
Note that  the equality only holds when the prior mixture degenerates to a factorized distribution with all mixture components equivalent.

\paragraph{Unbiased Stochastic Entropy and Prior Expectation Gradients}
We previously showed how to lower bound the ELBO terms
$\mat{q}^T\logp$ and $-\mat{q}^T\logq$ when the variational and/or prior distributions do not factor, however, optimizing this bound introduces bias and does not guarantee convergence to a local optimum of the true ELBO.
Here we reintroduce REINFORCE to deliver unbiased gradient estimates for these terms.
The REINFORCE estimator typically has high variance, however, since gradient estimates for these terms are so cheap, a massive number of samples can be used per stochastic gradient descent (SGD) iteration to decrease variance.
Since we can still compute the expensive $\mat{q}^T\logl$ term \emph{exactly} when $\mat{q}$ is an unfactorized mixture distribution, its gradient can be computed exactly.
The unbiased gradient estimator of $\mat{q}^T\logq$ is expressed as follows%
\footnote{We used the identity
$\big(\logq + 1\big) \odot \frac{\partial \logq}{\partial \theta} 
= \frac{1}{2}\frac{\partial }{\partial \theta}\big(\logq + 1\big)^2$, where $\odot$ denotes an elementwise product.}
\begin{align} \label{eqn:entropy_surrogate}
%\frac{\partial}{\partial \theta} \mat{q}^T\logp
%&= \mat{q}^T \Big(\logp \odot \frac{\partial \logq}{\partial \theta} + \frac{\partial \logp}{\partial \theta}\Big),\\
%&= \mat{q}^T \Big(\logp \odot \frac{\partial \logq}{\partial \theta}\Big),\\
\frac{\partial}{\partial \theta} \mat{q}^T\logq
%= \mat{q}^T \Big(\big(\logq + 1\big) \odot \frac{\partial \logq}{\partial \theta}\Big)
= \frac{1}{2}\mat{q}^T \bigg(\frac{\partial}{\partial \theta}\big(\logq + 1\big)^2\bigg)
\approx \frac{\partial}{\partial \theta} \frac{1}{2t}\sum_{i=1}^t \big(\log q(\mat{s}_i) + 1\big)^2,
\end{align}
where $\mat{s}_i \inR{b}$ is the $i$th of $t$ samples from the variational distribution used in the Monte Carlo gradient estimator.
It is evident that this surrogate loss can be easily optimized using automatic differentiation, and the per-sample computations are extremely cheap.

\section{Numerical Studies}
\label{sec:studies}
\subsection{Comparison with REINFORCE}
\begin{figure}
\centering
\includegraphics[width=\textwidth]{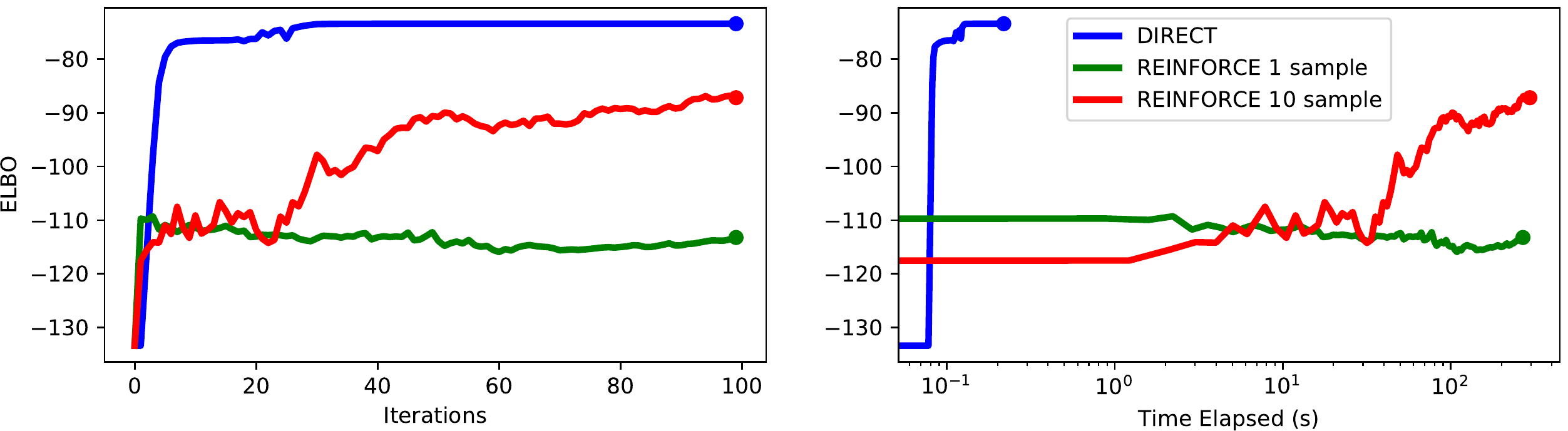}
\caption{Convergence rates of a GLM trained with REINFORCE verses the proposed DIRECT method. 
	The DIRECT method greatly outperforms REINFORCE in iterations and wall-clock time.}
\label{fig:reinforce_comparison}
\end{figure}

As discussed in \cref{sec:background}, we cannot reparameterize because of the discrete latent variable priors considered, however, we can directly compare the optimization performance of the proposed techniques with the REINFORCE gradient estimator~\cite{williams_reinforce}.
In \cref{fig:reinforce_comparison}, we compare ELBO maximization performance between the proposed DIRECT, and the REINFORCE methods.
For this study we generated a dataset from a random weighting of $b=20$ random Fourier features of a squared exponential kernel~\cite{rahimi_rff} and corrupted by independent Gaussian noise.
We use a generalized linear regression model as described in \cref{sec:glm} which uses the same features with $\widebar{m}=3$.
We consider a prior over $\sigma^2$,
and a mean-field variational distribution giving $\widebar{m}(b+1) = 63$ variational parameters which we initialize to be the same as the prior; a uniform categorical distribution.
For DIRECT, a L-BFGS optimizer is used~\cite{nocedal_lbfgs} and stochastic gradient descent is used for REINFORCE with a varying number of samples used for the Monte Carlo gradient estimator.
Both methods use full batch training and are implemented using TensorFlow~\cite{tensorflow}.
It can be seen that DIRECT greatly outperforms REINFORCE both in the number of iterations and computational time.
As we move to a large $n$ or a larger $b$, the difference between the proposed DIRECT technique and REINFORCE becomes more profound.
The superior scaling with respect to $n$ was expected since we had shown in \cref{sec:glm} that the DIRECT computational runtime is independent of $n$.
However, the improved scaling with respect to $b$ is an interesting result and may be attributed to the fact that as the dimension of the variational parameter space increases, there is more value in having low (or zero) variance estimates of the gradient.

\subsection{Relaxing Gaussian Priors on UCI Regression Datasets}
In this section, we consider discretely relaxing a continuous Gaussian prior on the weights of a generalized linear regression model.
This allows us to compare performance between a reparameterization gradient estimator for a continuous prior and our DIRECT method for a relaxed, discrete prior.

Considering regression datasets from the UCI repository, we report the mean and standard deviation of the root mean squared error (RMSE) from 10-fold cross validation%
\footnote{90\% train, 10\% test per fold. We use folds from \url{https://people.orie.cornell.edu/andrew/code}}.
Also presented is the mean training time per fold on a machine with two E5-2680 v3 processors and 128Gb of RAM, and the expected sparsity (percentage of zeros) within a posterior sample.
All models use $b=2000$ basis functions.
Further details of the experimental setup can be found in \cref{sec:uci_setup}.
In \cref{tbl:uci_results}, we see the results of our studies across several model-types.
In the left column, the ``REPARAM Mean-Field'' model uses a (continuous) Gaussian prior, an uncorrelated Gaussian variational distribution and reparameterization gradients.
The right two models use a discrete relaxation of a Gaussian prior~(DIRECT) with support at 15 discrete values, allowing storage of each latent variable sample as a vector of 4-bit quantized integers.
Therefore, each ELBO evaluation requires $15^{2000} > 10^{2352}$ log-likelihood evaluations, however, these computation can be done quickly by exploiting Kronecker matrix algebra.
We compute the ELBO as described in \cref{sec:glm} for the ``DIRECT Mean-Field'' model, and use the low-variance, unbiased gradient estimator described in \cref{eqn:entropy_surrogate} for the ``DIRECT 5-Mixture SGD'' model which uses a mixture distribution with $r=5$ components, and $t=3000$ Monte Carlo samples for the entropy gradient estimator.

The boldface entries indicate top performance on each dataset, where
it is evident that the DIRECT method not only outperformed REPARAM on most datasets but also trained much faster, particularly on the large datasets due to the independence of dataset size on computational complexity.
The DIRECT mean-field model contains $\widebar{m}b=30,000$ variational parameters, however,
training took just seconds on all datasets, including \emph{electric} with over 2~million points.
The DIRECT mixture model contains $\widebar{m}br=150,000$ variational parameters, and since the gradient estimates are stochastic, average training times are on the order of hundreds of seconds across all datasets.
While the time for precomputations does depend on dataset size, its contribution to the overall timings are negligible, being well under one second for the largest dataset, \emph{electric}.
Additionally, it is evident that posterior samples from the DIRECT model tend to be very sparse. 
For example, the DIRECT models on the \emph{gas} dataset admit posterior samples that are over 84\% sparse on average, meaning that over 1680 weights are expected to be zero in a posterior sample with $b=2000$ elements.
This would yield massive computational savings on hardware limited devices.
Samples from the DIRECT models on the \emph{electric} dataset are over 99.6\% sparse.
% which may suggest that many basis functions had poor alignment with the targets and were pruned from the model.

Comparing the DIRECT mean-field model to the mixture model, we observe gains in the RMSE performance on many datasets, as we would expect with the increased flexibility of the variational distribution.
While we only showed the posterior mean in our results, we would expect an even larger disparity in the quality of the predictive uncertainty which was not analyzed.
In \cref{tbl:uci_biased_results} of the supplement, we present results for a DIRECT mixture model that uses the ELBO lower bound presented in \Cref{thm:mixture_lb}.
This model does not perform as well as the DIRECT mixture model trained using an unbiased SGD approach, as would be expected, however, 
it does train faster since its objective is evaluated deterministically, and
its RMSE performance is still marginally better than the DIRECT mean-field model on many datasets.

\begin{table}
	\centering
	\setlength\tabcolsep{1.5pt} % default value: 6pt
	\scriptsize%\hspace{-1cm}%
	\begin{tabular}{@{}|lll|lll|lll||ll|@{}}\toprule%&&&&&&&&&&&
&&&\multicolumn{3}{c|}{Continuous Prior}&\multicolumn{5}{c|}{Discrete 4-bit Prior}\\%&\\%&&&&&&
&&&\multicolumn{3}{c|}{REPARAM Mean-Field}&\multicolumn{3}{c||}{DIRECT Mean-Field}&\multicolumn{2}{c|}{DIRECT 5-Mixture SGD}\\%&\\%&&&&&
Dataset&$n$&$d$&Time&RMSE&Sparsity&Time&RMSE&Sparsity&RMSE&Sparsity\\%&\\%
\midrule%&&&&&&&&&&&
challenger&23&4&8&$\bf 0.515 \pm 0.284$&0\%&1&$0.523 \pm 0.248$&17\%&$0.525 \pm 0.246$&17\%\\%&\\%
fertility&100&9&8&$0.161 \pm 0.043$&0\%&2&$\bf 0.159 \pm 0.041$&17\%&$0.16 \pm 0.041$&17\%\\%&\\%
%slump&103&7&8&$754.237 \pm 588.176$&0\%&1&$56.662 \pm 13.351$&99.8\%&$\bf 54.588 \pm 11.993$&99.7\%\\%&\\%
automobile&159&25&5&$0.425 \pm 0.2$&0\%&10&$0.129 \pm 0.063$&51\%&$\bf 0.122 \pm 0.056$&51\%\\%&\\%
servo&167&4&5&$0.524 \pm 0.184$&0\%&10&$\bf 0.271 \pm 0.08$&35\%&$0.274 \pm 0.077$&35\%\\%&\\%
cancer&194&33&5&$27.488 \pm 5.45$&0\%&4&$22.954 \pm 3.09$&19\%&$\bf 22.937 \pm 3.135$&19\%\\%&\\%
hardware&209&7&5&$1.796 \pm 1.537$&0\%&11&$\bf 0.401 \pm 0.048$&51\%&$\bf 0.401 \pm 0.046$&51\%\\%&\\%
yacht&308&6&5&$0.815 \pm 0.17$&0\%&1&$0.234 \pm 0.07$&96\%&$\bf 0.225 \pm 0.082$&96\%\\%&\\%
autompg&392&7&5&$4.05 \pm 0.739$&0\%&10&$2.564 \pm 0.363$&31\%&$\bf 2.543 \pm 0.362$&31\%\\%&\\%
housing&506&13&5&$3.014 \pm 0.567$&0\%&10&$\bf 2.752 \pm 0.405$&40\%&$\bf 2.699 \pm 0.361$&39\%\\%&\\%
forest&517&12&5&$1.378 \pm 0.148$&0\%&2&$1.363 \pm 0.15$&17\%&$\bf 1.357 \pm 0.155$&17\%\\%&\\%
stock&536&11&5&$0.751 \pm 0.338$&0\%&8&$0.011 \pm 0.003$&98\%&$\bf 0.008 \pm 0.001$&98\%\\%&\\%
pendulum&630&9&5&$1.465 \pm 0.26$&0\%&1&$1.329 \pm 0.282$&68\%&$\bf 1.312 \pm 0.253$&63\%\\%&\\%
energy&768&8&5&$78.852 \pm 21.73$&0\%&1&$3.272 \pm 0.332$&99\%&$\bf 2.911 \pm 0.309$&99\%\\%&\\%
concrete&1030&8&5&$10.347 \pm 2.847$&0\%&10&$\bf 5.316 \pm 0.716$&82\%&$5.477 \pm 0.632$&82\%\\%&\\%
solar&1066&10&5&$0.902 \pm 0.171$&0\%&10&$\bf 0.787 \pm 0.192$&23\%&$0.788 \pm 0.189$&23\%\\%&\\%
airfoil&1503&5&5&$\bf 2.071 \pm 0.271$&0\%&11&$2.175 \pm 0.349$&48\%&$2.156 \pm 0.316$&45\%\\%&\\%
wine&1599&11&5&$0.939 \pm 0.33$&0\%&11&$0.472 \pm 0.044$&54\%&$\bf 0.469 \pm 0.042$&54\%\\%&\\%
gas&2565&128&5&$0.27 \pm 0.052$&0\%&1&$0.211 \pm 0.058$&84\%&$\bf 0.184 \pm 0.063$&76\%\\%&\\%
skillcraft&3338&19&46&$0.273 \pm 0.029$&0\%&7&$\bf 0.253 \pm 0.016$&97\%&$\bf 0.253 \pm 0.016$&97\%\\%&\\%
sml&4137&26&47&$\bf 0.327 \pm 0.013$&0\%&1&$0.677 \pm 0.044$&57\%&$0.671 \pm 0.047$&57\%\\%&\\%
parkinsons&5875&20&48&$\bf 0.158 \pm 0.009$&0\%&1&$0.651 \pm 0.034$&13\%&$0.613 \pm 0.083$&13\%\\%&\\%
%pumadyn&8192&32&48&$\bf 0.609 \pm 0.224$&0\%&1&$0.997 \pm 0.031$&100\%&$0.937 \pm 0.069$&34\%\\%&\\%
poletele&15000&26&50&$\bf 12.487 \pm 0.363$&0\%&10&$13.65 \pm 0.348$&16\%&$13.369 \pm 0.431$&17\%\\%&\\%
elevators&16599&18&51&$0.247 \pm 0.156$&0\%&1&$\bf 0.124 \pm 0.003$&99\%&$\bf 0.124 \pm 0.003$&99\%\\%&\\%
%kin40k&40000&8&56&$\bf 0.272 \pm 0.009$&0\%&2&$0.408 \pm 0.017$&70\%&$0.343 \pm 0.014$&62\%\\%&\\%
protein&45730&9&58&$0.642 \pm 0.006$&0\%&11&$0.619 \pm 0.007$&76\%&$\bf 0.618 \pm 0.007$&60\%\\%&\\%
kegg&48827&20&58&$\bf 0.178 \pm 0.012$&0\%&1&$0.222 \pm 0.009$&96\%&$0.205 \pm 0.004$&95\%\\%&\\%
ctslice&53500&385&61&$\bf 4.415 \pm 0.113$&0\%&2&$6.063 \pm 0.122$&19\%&$5.478 \pm 0.137$&42\%\\%&\\%
keggu&63608&27&61&$\bf 0.122 \pm 0.004$&0\%&1&$0.139 \pm 0.004$&87\%&$0.136 \pm 0.006$&87\%\\%&\\%
3droad&434874&3&141&$11.057 \pm 0.091$&0\%&2&$10.493 \pm 0.105$&40\%&$\bf 10.354 \pm 0.077$&33\%\\%&\\%
song&515345&90&158&$0.537 \pm 0.002$&0\%&2&$ 0.501 \pm 0.002$&32\%&$\bf 0.498 \pm 0.002$&28\%\\%&\\%
buzz&583250&77&169&$\bf 0.94 \pm 0.006$&0\%&1&$1.007 \pm 0.007$&82\%&$0.959 \pm 0.004$&80\%\\%&\\%
electric&2049280&11&500&$9.26 \pm 4.47$&0\%&1&$0.575 \pm 0.032$&99.6\%&$\bf 0.557 \pm 0.055$&99.6\%\\%&\\%
\bottomrule \end{tabular}%&&&&&&&&&&&

	\caption{Mean and standard deviation of test error, average training time, and average expected sparsity of a posterior sample from 10-fold cross validation on UCI regression datasets.}
	\label{tbl:uci_results}
\end{table}

\section{Conclusions}
\label{sec:conclusions}
We have shown that by discretely relaxing continuous priors, variational inference can be performed accurately and efficiently using our DIRECT method.
We have demonstrated that through the use of Kronecker matrix algebra, the ELBO of a discretely relaxed model can be efficiently and exactly computed even when this computation requires significantly more log-likelihood evaluations than the number of atoms in the known universe.
Through this ability to exactly perform ELBO computations we achieve unbiased, zero-variance gradient estimates using automatic differentiation which we show significantly outperforms competing Monte Carlo alternatives that admit high-variance gradient estimates.
We also demonstrate that the computational complexity of ELBO computations is \emph{independent} of the quantity of training data using the DIRECT method, making the proposed approaches amenable to big data applications.
At inference time, we show that we can again use Kronecker matrix algebra to exactly compute the statistical moments of the parameterized predictive posterior distribution, unlike competing techniques which rely on Monte Carlo sampling.
Finally, we discuss and demonstrate how posterior samples can be sparse and can be represented as quantized integer values to enable efficient inference which is particularly powerful on hardware limited devices, 
or if energy efficiency is a major concern.

We illustrate the DIRECT approach on several popular models such as mean-field variational inference for generalized linear models and deep Bayesian neural networks for regression.
We also discuss some models which do not admit a compact representation for exact ELBO computations.
For these cases, we discuss and demonstrate novel extensions to the DIRECT method that allow efficient computation of a lower bound of the ELBO, and we demonstrate how an unfactorized variational distribution can be used by introducing a manageable level of stochasticity into the gradients.
We hope that these new approaches for ELBO computations will inspire new model structures and research directions in approximate Bayesian inference.

\newpage
{
%\small
\section*{Acknowledgements} Research funded by an NSERC Discovery Grant and the Canada Research Chairs~program.
}
\bibliographystyle{IEEEtranN}
\bibliography{peripheral,core,references}

% Generated by IEEEtranN.bst, version: 1.13 (2008/09/30)
\begin{thebibliography}{37}
\providecommand{\natexlab}[1]{#1}
\providecommand{\url}[1]{#1}
\csname url@samestyle\endcsname
\providecommand{\newblock}{\relax}
\providecommand{\bibinfo}[2]{#2}
\providecommand{\BIBentrySTDinterwordspacing}{\spaceskip=0pt\relax}
\providecommand{\BIBentryALTinterwordstretchfactor}{4}
\providecommand{\BIBentryALTinterwordspacing}{\spaceskip=\fontdimen2\font plus
\BIBentryALTinterwordstretchfactor\fontdimen3\font minus
  \fontdimen4\font\relax}
\providecommand{\BIBforeignlanguage}[2]{{%
\expandafter\ifx\csname l@#1\endcsname\relax
\typeout{** WARNING: IEEEtranN.bst: No hyphenation pattern has been}%
\typeout{** loaded for the language `#1'. Using the pattern for}%
\typeout{** the default language instead.}%
\else
\language=\csname l@#1\endcsname
\fi
#2}}
\providecommand{\BIBdecl}{\relax}
\BIBdecl

\bibitem[Thrun et~al.(2005)Thrun, Burgard, and
  Fox]{thrun_probabilistic_robotics}
S.~Thrun, W.~Burgard, and D.~Fox, \emph{Probabilistic robotics}.\hskip 1em plus
  0.5em minus 0.4em\relax MIT press, 2005.

\bibitem[Bradshaw et~al.(2017)Bradshaw, Matthews, and
  Ghahramani]{ghahramani_adversarial}
J.~Bradshaw, A.~G. d.~G. Matthews, and Z.~Ghahramani, ``Adversarial examples,
  uncertainty, and transfer testing robustness in gaussian process hybrid deep
  networks,'' Tech. Rep., 2017.

\bibitem[Louizos et~al.(2017)Louizos, Ullrich, and
  Welling]{welling_bayes_compression}
C.~Louizos, K.~Ullrich, and M.~Welling, ``Bayesian compression for deep
  learning,'' in \emph{Advances in Neural Information Processing Systems},
  2017, pp. 3288--3298.

\bibitem[Jordan et~al.(1999)Jordan, Ghahramani, Jaakkola, and
  Saul]{jordan_vi_root}
M.~I. Jordan, Z.~Ghahramani, T.~S. Jaakkola, and L.~K. Saul, ``An introduction
  to variational methods for graphical models,'' \emph{Machine learning},
  vol.~37, no.~2, pp. 183--233, 1999.

\bibitem[Wainwright and Jordan(2008)]{wainwright_vi_root}
M.~J. Wainwright and M.~I. Jordan, ``Graphical models, exponential families,
  and variational inference,'' \emph{Foundations and Trends in Machine
  Learning}, vol.~1, no. 1--2, pp. 1--305, 2008.

\bibitem[Hoffman et~al.(2013)Hoffman, Blei, Wang, and Paisley]{hoffman_svi}
M.~D. Hoffman, D.~M. Blei, C.~Wang, and J.~Paisley, ``Stochastic variational
  inference,'' \emph{The Journal of Machine Learning Research}, vol.~14, no.~1,
  pp. 1303--1347, 2013.

\bibitem[Ranganath et~al.(2014)Ranganath, Gerrish, and
  Blei]{ranganath_blackbox_vi}
R.~Ranganath, S.~Gerrish, and D.~Blei, ``Black box variational inference,'' in
  \emph{Artificial Intelligence and Statistics}, 2014, pp. 814--822.

\bibitem[Kucukelbir et~al.(2017)Kucukelbir, Tran, Ranganath, Gelman, and
  Blei]{blei_blackbox_vi}
A.~Kucukelbir, D.~Tran, R.~Ranganath, A.~Gelman, and D.~M. Blei, ``Automatic
  differentiation variational inference,'' \emph{The Journal of Machine
  Learning Research}, vol.~18, no.~1, pp. 430--474, 2017.

\bibitem[Blei et~al.(2017)Blei, Kucukelbir, and McAuliffe]{blei_vi_review}
D.~M. Blei, A.~Kucukelbir, and J.~D. McAuliffe, ``Variational inference: A
  review for statisticians,'' \emph{Journal of the American Statistical
  Association}, vol. 112, no. 518, pp. 859--877, 2017.

\bibitem[Kingma and Welling(2013)]{kingma_reparam}
D.~P. Kingma and M.~Welling, ``Auto-encoding variational {B}ayes,'' \emph{arXiv
  preprint arXiv:1312.6114}, 2013.

\bibitem[Williams(1992)]{williams_reinforce}
R.~J. Williams, ``Simple statistical gradient-following algorithms for
  connectionist reinforcement learning,'' in \emph{Reinforcement Learning},
  1992, pp. 5--32.

\bibitem[Fu(2006)]{fu_score_function}
M.~C. Fu, ``Gradient estimation,'' \emph{Handbooks in operations research and
  management science}, vol.~13, pp. 575--616, 2006.

\bibitem[Glynn(1990)]{glynn_likelihood_ratio}
P.~W. Glynn, ``Likelihood ratio gradient estimation for stochastic systems,''
  \emph{Communications of the ACM}, vol.~33, no.~10, pp. 75--84, 1990.

\bibitem[Jang et~al.(2016)Jang, Gu, and Poole]{jang_gumbel_softmax}
E.~Jang, S.~Gu, and B.~Poole, ``Categorical reparameterization with
  {G}umbel-softmax,'' \emph{arXiv preprint arXiv:1611.01144}, 2016.

\bibitem[Maddison et~al.(2016)Maddison, Mnih, and Teh]{maddison_concrete}
C.~J. Maddison, A.~Mnih, and Y.~W. Teh, ``The concrete distribution: A
  continuous relaxation of discrete random variables,'' \emph{arXiv preprint
  arXiv:1611.00712}, 2016.

\bibitem[Tucker et~al.(2017)Tucker, Mnih, Maddison, Lawson, and
  Sohl-Dickstein]{tucker_rebar}
G.~Tucker, A.~Mnih, C.~J. Maddison, J.~Lawson, and J.~Sohl-Dickstein,
  ``{REBAR}: Low-variance, unbiased gradient estimates for discrete latent
  variable models,'' in \emph{Advances in Neural Information Processing
  Systems}, 2017, pp. 2627--2636.

\bibitem[Grathwohl et~al.(2017)Grathwohl, Choi, Wu, Roeder, and
  Duvenaud]{grathwohl_relax}
W.~Grathwohl, D.~Choi, Y.~Wu, G.~Roeder, and D.~Duvenaud, ``Backpropagation
  through the void: Optimizing control variates for black-box gradient
  estimation,'' in \emph{International Conference on Learning Representations},
  2017.

\bibitem[Van~Loan(2000)]{van_loan_kron}
C.~F. Van~Loan, ``The ubiquitous {K}ronecker product,'' \emph{Journal of
  Computational and Applied Mathematics}, vol. 123, no.~1, pp. 85--100, 2000.

\bibitem[Horn and Johnson(1994)]{kron_sum}
R.~A. Horn and C.~R. Johnson, \emph{Topics in Matrix analysis}.\hskip 1em plus
  0.5em minus 0.4em\relax Cambridge university press, 1994.

\bibitem[Wilson et~al.(2016)Wilson, Hu, Salakhutdinov, and Xing]{wilson_deep}
A.~G. Wilson, Z.~Hu, R.~Salakhutdinov, and E.~P. Xing, ``Deep kernel
  learning,'' in \emph{Artificial Intelligence and Statistics}, 2016, pp.
  370--378.

\bibitem[Chen et~al.(2015)Chen, Wilson, Tyree, Weinberger, and
  Chen]{chen_compressing_nn}
W.~Chen, J.~Wilson, S.~Tyree, K.~Weinberger, and Y.~Chen, ``Compressing neural
  networks with the hashing trick,'' in \emph{International Conference on
  Machine Learning}, 2015, pp. 2285--2294.

\bibitem[Gong et~al.(2014)Gong, Liu, Yang, and Bourdev]{gong_compressing_nn}
Y.~Gong, L.~Liu, M.~Yang, and L.~Bourdev, ``Compressing deep convolutional
  networks using vector quantization,'' \emph{arXiv preprint arXiv:1412.6115},
  2014.

\bibitem[Han et~al.(2015)Han, Mao, and Dally]{han_compressing_nn}
S.~Han, H.~Mao, and W.~J. Dally, ``Deep compression: Compressing deep neural
  networks with pruning, trained quantization and huffman coding,'' \emph{arXiv
  preprint arXiv:1510.00149}, 2015.

\bibitem[Zhou et~al.(2017)Zhou, Yao, Guo, Xu, and Chen]{zhou_compression_nn}
A.~Zhou, A.~Yao, Y.~Guo, L.~Xu, and Y.~Chen, ``Incremental network
  quantization: Towards lossless cnns with low-precision weights,'' \emph{arXiv
  preprint arXiv:1702.03044}, 2017.

\bibitem[Hubara et~al.(2016)Hubara, Courbariaux, Soudry, El-Yaniv, and
  Bengio]{bengio_binary_nn}
I.~Hubara, M.~Courbariaux, D.~Soudry, R.~El-Yaniv, and Y.~Bengio, ``Binarized
  neural networks,'' in \emph{Advances in neural information processing
  systems}, 2016, pp. 4107--4115.

\bibitem[Li et~al.(2016)Li, Zhang, and Liu]{li_ternary_nn}
F.~Li, B.~Zhang, and B.~Liu, ``Ternary weight networks,'' \emph{arXiv preprint
  arXiv:1605.04711}, 2016.

\bibitem[Rastegari et~al.(2016)Rastegari, Ordonez, Redmon, and
  Farhadi]{rastegari_xnor_nn}
M.~Rastegari, V.~Ordonez, J.~Redmon, and A.~Farhadi, ``Xnor-net: Imagenet
  classification using binary convolutional neural networks,'' in
  \emph{European Conference on Computer Vision}.\hskip 1em plus 0.5em minus
  0.4em\relax Springer, 2016, pp. 525--542.

\bibitem[Jacob et~al.(2017)Jacob, Kligys, Chen, Zhu, Tang, Howard, Adam, and
  Kalenichenko]{jacob_quantization_nn}
B.~Jacob, S.~Kligys, B.~Chen, M.~Zhu, M.~Tang, A.~Howard, H.~Adam, and
  D.~Kalenichenko, ``Quantization and training of neural networks for efficient
  integer-arithmetic-only inference,'' \emph{arXiv preprint arXiv:1712.05877},
  2017.

\bibitem[Lazarsfeld and Henry(1968)]{lazarsfeld_lsa}
P.~Lazarsfeld and N.~Henry, \emph{Latent structure analysis}.\hskip 1em plus
  0.5em minus 0.4em\relax Houghton Mifflin Company, Boston, Massachusetts,
  1968.

\bibitem[Goodman(1974)]{goodman_lsa}
L.~A. Goodman, ``Exploratory latent structure analysis using both identifiable
  and unidentifiable models,'' \emph{Biometrika}, vol.~61, no.~2, pp. 215--231,
  1974.

\bibitem[Rahimi and Recht(2007)]{rahimi_rff}
A.~Rahimi and B.~Recht, ``Random features for large-scale kernel machines,'' in
  \emph{Advances in Neural Information Processing Systems}, 2007, pp.
  1177--1184.

\bibitem[Byrd et~al.(1995)Byrd, Lu, Nocedal, and Zhu]{nocedal_lbfgs}
R.~H. Byrd, P.~Lu, J.~Nocedal, and C.~Zhu, ``A limited memory algorithm for
  bound constrained optimization,'' \emph{SIAM Journal on Scientific
  Computing}, vol.~16, no.~5, pp. 1190--1208, 1995.

\bibitem[Abadi et~al.(2016)Abadi, Barham, Chen, Chen, Davis, Dean, Devin,
  Ghemawat, Irving, Isard, et~al.]{tensorflow}
M.~Abadi, P.~Barham, J.~Chen, Z.~Chen, A.~Davis, J.~Dean, M.~Devin,
  S.~Ghemawat, G.~Irving, M.~Isard \emph{et~al.}, ``{T}ensor{F}low: A system
  for large-scale machine learning.'' in \emph{OSDI}, vol.~16, 2016, pp.
  265--283.

\bibitem[Bishop(2006)]{bishop_ml}
C.~M. Bishop, \emph{Pattern Recognition and Machine Learning}.\hskip 1em plus
  0.5em minus 0.4em\relax Springer, 2006.

\bibitem[Nielsen and Sun(2016)]{logsumexp}
F.~Nielsen and K.~Sun, ``Guaranteed bounds on the {K}ullback-{L}eibler
  divergence of univariate mixtures using piecewise log-sum-exp inequalities,''
  \emph{arXiv:1606.05850}, 2016.

\bibitem[Rasmussen and Williams(2006)]{rasmussen_gpml}
C.~E. Rasmussen and C.~K.~I. Williams, \emph{Gaussian Processes for Machine
  Learning}.\hskip 1em plus 0.5em minus 0.4em\relax MIT Press, 2006.

\bibitem[Tran et~al.(2016)Tran, Kucukelbir, Dieng, Rudolph, Liang, and
  Blei]{edward}
D.~Tran, A.~Kucukelbir, A.~B. Dieng, M.~Rudolph, D.~Liang, and D.~M. Blei,
  ``{Edward: A library for probabilistic modeling, inference, and criticism},''
  \emph{arXiv preprint arXiv:1610.09787}, 2016.

\end{thebibliography}
\appendix

\newpage
\section{Proof of \Cref{thm:glm}: ELBO Computation for a Regression GLM}
\label{sec:glm_proof}
For our generalized linear regression model with a prior over $\sigma^2$, we can re-write \cref{eqn:ELBO2} as follows
\begin{align} \label{eqn:ELBO_w_sigma}
\textsf{ELBO}(\mat{\theta})
= (\mat{q}_{\sigma} \otimes \mat{q})^T \logl 
+ \sum_{i=1}^b \mat{q}_i^T \logp_i 
- \sum_{i=1}^b \mat{q}_i^T \logq_i
+ \mat{q}_{\sigma}^T \logp_{\sigma}
- \mat{q}_{\sigma}^T \logq_{\sigma},
\end{align}
where we have simply expanded the factorized variational distribution to include $\sigma^2$, resulting in the two extra terms.
To complete the ELBO in \cref{eqn:ELBO_w_sigma}, we need to take the inner product between the variational distribution and log-likelihood for each point in the hypothesis space, $(\mat{q}_{\sigma} \otimes \mat{q})^T \logl$. 
We can write this relation as follows for our generalized linear regression model,~(see e.g.~\cite{bishop_ml})
\begin{align} \label{eqn:glm_qTlogl}
%p(\mat{y} | \mat{X}, \mat{w}) &= (\sigma^2)^{-\frac{n}{2}} \exp \bigg(
%-\frac{1}{2\sigma^2} (\mat{y} - \mPhi \mat{w})^T (\mat{y} - \mPhi \mat{w}) \bigg),\\
(\mat{q}_{\sigma} \otimes \mat{q})^T \logl =
-\frac{n}{2}\mat{q}_{\sigma}^T \log\mat{\sigma}^2
-\frac{1}{2}\big(\mat{q}_{\sigma}^T \mat{\sigma}^{-2}\big) \big(\mat{q}^T \{(\mat{y} - \mPhi \mat{w}_i)^T (\mat{y} - \mPhi \mat{w}_i)\}_{i=1}^m \big),
\end{align}
whose computation would be prohibitively expensive when $m=\widebar{m}^b$ is large.
We will now focus on computing the inner product involving the variational distribution over the $\mat{w}$ variables, $\mat{q}$, which we can break into three terms as follows,
\begin{align}
\mat{q}^T\{(\mat{y} - \mPhi \mat{w}_i)^T (\mat{y} - \mPhi \mat{w}_i)\}_{i=1}^m
&= \mat{y}^T\mat{y} 
- 2\mat{q}^T\{\mat{y}^T \mat{\Phi} \mat{w}_i\}_{i=1}^m 
+ \mat{q}^T\{\mat{w}_i^T \mat{\Phi}^T \mat{\Phi} \mat{w}_i\}_{i=1}^m, \label{eqn:glm_terms}
%&= \sum_{i=1}^n y_i^2 - \sum_{i=1}^n 2 y_i \mat{\Phi}[i,:] \mat{w} + \sum_{i=1}^n(\mat{\Phi}[i,:] \mat{w})^2 
\end{align}
for which the first term is trivial to compute as written since it does not depend on $\mat{w}$.
We now demonstrate how the second and third terms can be computed, recalling that we have assumed that $\mat{q}$ is a mean-field variational distribution.
Firstly, define 
$\mat{Z} = (\mat{\Phi}\W)^T = \{\oplus_{j=1}^b \phi_{ij} \widebar{\mat{w}}_j\}_{i=1}^n \inR{m \times n}$ whose columns contain the model prediction for a single training point at every possible set of latent variable values in the hypothesis space.
Observe that each column is represented as a sum of $b$ Kronecker product vectors.
%$\mat{z}_i = (\mat{\Phi}(i,:)\W)^T = \oplus_{j=1}^b \phi_{ij} \widebar{\mat{w}}_j \inR{m}$
We can then write the second term of \cref{eqn:glm_terms} as
\begin{align} \label{eqn:glm_2}
\hspace{-3mm}
\mat{q}^T \{\mat{y}^T \mat{\Phi} \mat{w}_i\}_{i=1}^m =
\sum_{i=1}^n y_i\mat{q}^T\mat{z}_i 
&= \sum_{i=1}^n y_i\sum_{j=1}^b \phi_{ij}\mat{q}_j^T\widebar{\mat{w}}_j
= \sum_{j=1}^b  \mat{q}_j^T\widebar{\mat{w}}_j \bigg(\sum_{i=1}^n y_i\phi_{ij}\bigg)
= \mat{s}^T \big(\mat{\Phi}^T \mat{y}\big),
% this is validated in `validating_little_computations.ipynb`
\end{align}
where 
$\mat{s} = \{\mat{q}_j^T\widebar{\mat{w}}_j\}_{j=1}^b \inR{b}$.
%$\mat{s} = (\mat{Q}^T \odot \widebar{\mat{W}}^T)\mat{1}_{\widebar{m}} \inR{b}$, 
%$\mat{Q} = \{\mat{q}_i\}_{i=1}^b \inR{\widebar{m} \times b}$, and
%$\widebar{\mat{W}} = \{\widebar{\mat{w}}_i\}_{i=1}^b \inR{\widebar{m} \times b}$.
Finally, considering the third term of \cref{eqn:glm_terms}, observe that we can write
$\{\mat{w}_i^T \mat{\Phi}^T \mat{\Phi} \mat{w}_i\}_{i=1}^m = \sum_{i=1}^n \mat{z}_i^2$, and since each $\mat{z}_i$ is a sum of $b$ Kronecker product vectors, $\mat{z}_i^2$ a sum of $(b+b^2)/2$ Kronecker product vectors.
We can then write the third term of \cref{eqn:glm_terms} as follows,
\begin{align}
\mat{q}^T\{\mat{w}_i^T \mat{\Phi}^T \mat{\Phi} \mat{w}_i\}_{i=1}^m
&=
\sum_{i=1}^n \sum_{j=1}^{b} 
\mat{q}_j^T \widebar{\mat{w}}_j^2 \phi_{ij}^2
+ 2\sum_{k=j+1}^b \phi_{ij} \phi_{ik} 
(\mat{q}_{j}^T \widebar{\mat{w}}_{j})
(\mat{q}_{k}^T \widebar{\mat{w}}_{k}) ,\\
&= \sum_{j=1}^{b}\mat{q}_j^T \bigg(\widebar{\mat{w}}_j^2 \sum_{i=1}^n \phi_{ij}^2\bigg) 
+ 2\sum_{k=j+1}^b s_{j} s_{k} \bigg(\sum_{i=1}^n \phi_{ij} \phi_{ik}\bigg),
\\ \label{eqn:glm_3}
&=
\mat{s}^T\mat{\Phi}^T\mat{\Phi}\mat{s} - \texttt{diag}(\mat{\Phi}^T\mat{\Phi})^T\mat{s}^2 + \sum_{j=1}^{b}  \mat{q}_j^T \mat{h}_j,
\end{align}
where we have used the short-hand notation
$\mat{H} = \{\widebar{\mat{w}}_j^2 \sum_{i=1}^n \phi_{ij}^2\}_{j=1}^b \inR{\widebar{m} \times b}$.
Substituting \cref{eqn:glm_2} and \cref{eqn:glm_3} into \cref{eqn:glm_terms}, we can re-write the inner product between the variational distribution and the log-likelihood in \cref{eqn:glm_qTlogl} as follows,
\begin{multline}
(\mat{q}_{\sigma} \otimes \mat{q})^T \logl
=-\frac{n}{2}\mat{q}_{\sigma}^T \log\mat{\sigma}^2
-\frac{1}{2}\big(\mat{q}_{\sigma}^T \mat{\sigma}^{-2}\big)
\Big( 
\mat{y}^T\mat{y}
- 2 \mat{s}^T \big(\mat{\Phi}^T \mat{y}\big)
+ \mat{s}^T\mat{\Phi}^T\mat{\Phi}\mat{s} -\\ \textsf{diag}(\mat{\Phi}^T\mat{\Phi})^T\mat{s}^2 + \sum_{j=1}^{b}  \mat{q}_j^T \mat{h}_j
\Big),
\end{multline}
and substituting this into \cref{eqn:ELBO_w_sigma} completes the proof. \hfill $\square$

\section{Proof of \Cref{thm:logistic}: Logistic Regression ELBO Lower Bound}
\label{sec:logistic_proof}
\begin{figure}
	\begin{minipage}[t]{0.48\textwidth}
		\centering
		\includegraphics[width=\textwidth]{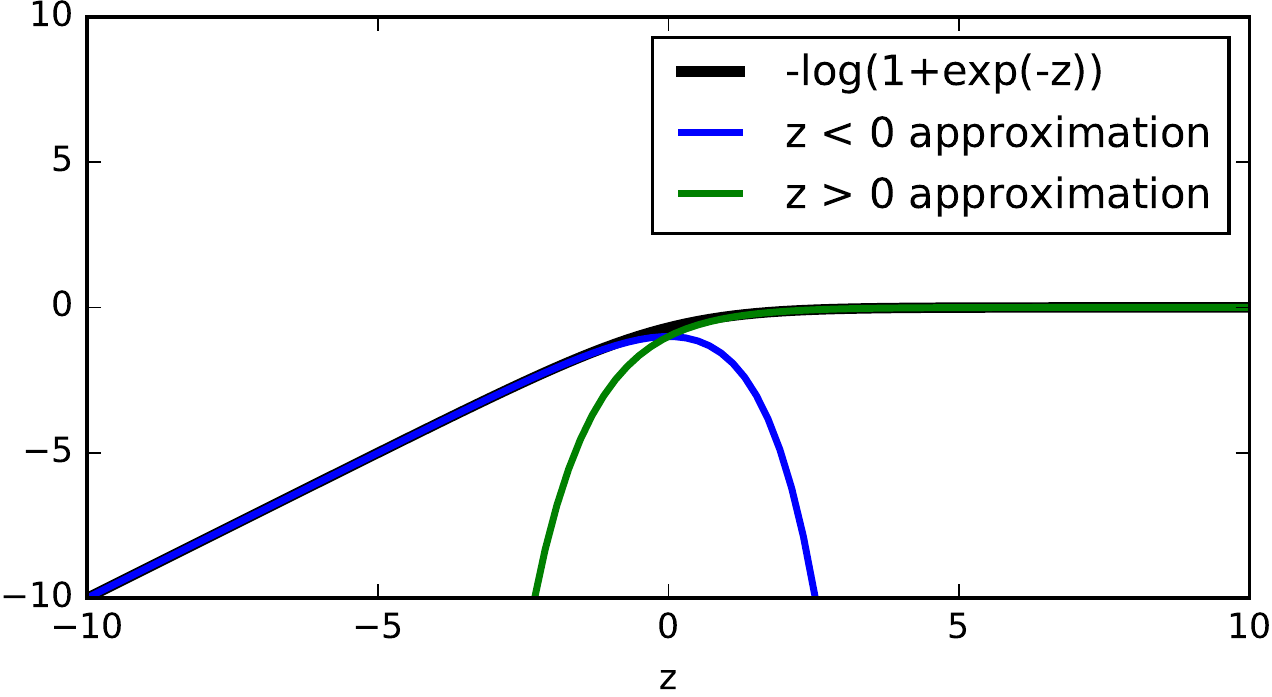}
		\caption{First-order Taylor series approximation of $-\log\big(1+\exp(-z)\big)$.
			The approximations evidently lower-bound the exact computation.}
		\label{fig:logistic_approx}
	\end{minipage}%
	\hspace{5mm}%
	\begin{minipage}[t]{0.48\textwidth}
		\centering
		\includegraphics[width=\textwidth]{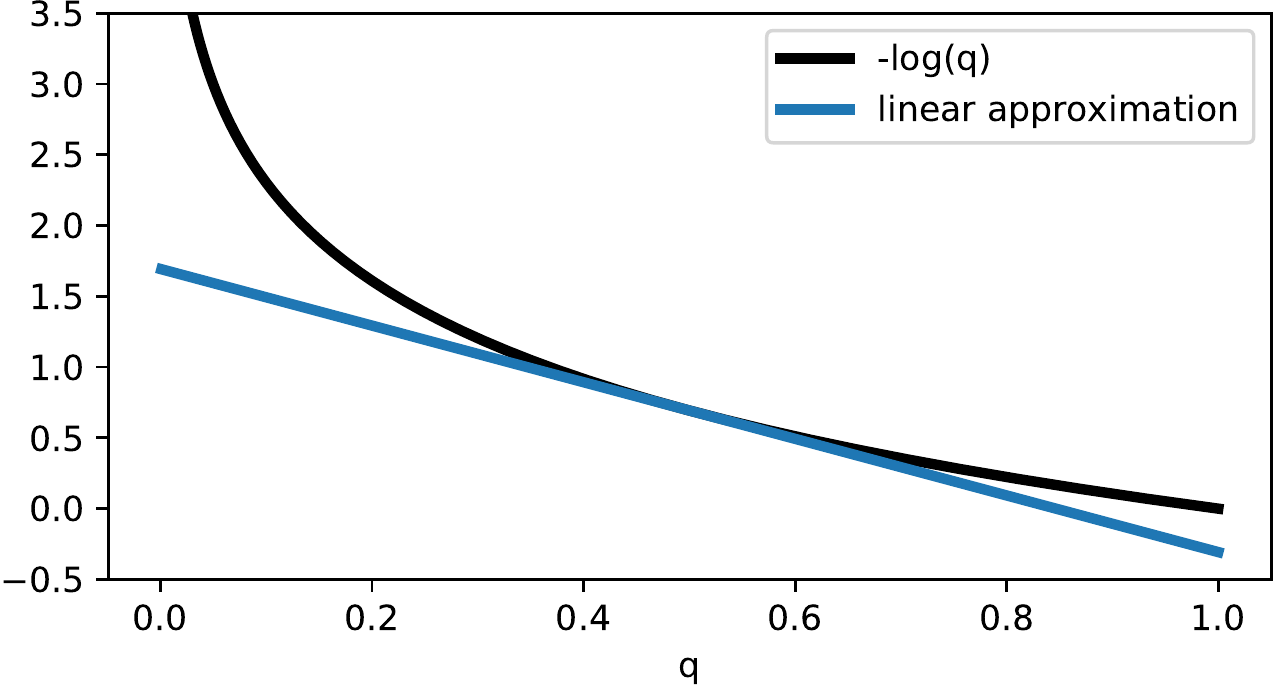}
		\caption{Taylor series approximation of $-\log(q)$ about $a=0.5$. The approximations evidently lower-bound the exact computation.}
		\label{fig:log_approx}
	\end{minipage}
\end{figure}

For the generalized linear logistic regression model considered, we can write the log likelihood as follows~(see e.g.~\cite{bishop_ml})
\begin{align}
\logl = \sum_{i=1}^n - y_i \mat{z}_i - \log\big(1+\exp(-\mat{z}_i)\big),
\end{align}
where $\mat{Z} = (\mat{\Phi}\W)^T = \{\oplus_{j=1}^b \phi_{ij} \widebar{\mat{w}}_j\}_{i=1}^n \inR{m \times n}$ is a matrix whose columns contain the logit values for a single training point at every possible set of latent variables in the hypothesis space.
It is evident that the first term is identical to that discussed in \cref{eqn:glm_2}, however, computation of the second term requires more development.
We can write
\begin{align}
\mat{q}^T\logl = 
-\mat{s}^T \big(\mat{\Phi}^T \mat{y}\big) 
-\sum_{i=1}^n \mat{q}^T \log\big(1+\exp(-\mat{z}_i)\big).
\end{align}
Since $\mat{z}_i = \oplus_{j=1}^b \phi_{ij} \widebar{\mat{w}}_j \inR{m}$ is a sum of $b$ Kronecker product vectors, each with one unique sub-matrix that is not unity, $\exp(-\mat{z}_i)$ is a single Kronecker product vector.
This follows from \Cref{prop:kron_sum}.
We can then take a Taylor series explanation of $\log\big(1+\exp(-\mat{z}_i)\big)$ as follows
\begin{align}
% https://www.wolframalpha.com/input/?i=log(1%2Bx)
\log(1 + \exp(-\mat{z}_i)) &= - \sum_{k=1}^\infty \frac{(-1)^k \exp(-k\mat{z}_i)}{k} &\text{ for } |\exp(-\mat{z}_i)|<1 \rightarrow \mat{z}_i > 0,\\
% accuracy plot: https://www.wolframalpha.com/input/?i=x+-+x%5E2%2F2+%2B+x%5E3%2F3+-+x%5E4%2F4+%2B+x%5E5%2F5+-+x%5E6%2F6+%2B+x%5E7%2F7+-+x%5E8%2F8+%2B+x%5E9%2F9+-+x%5E10%2F10+-+log(1%2Bx)+from+-1+to+2
\log(1 + \exp(-\mat{z}_i)) &= -\mat{z}_i - \sum_{k=1}^\infty \frac{(-1)^k \exp(k\mat{z}_i)}{k} &\text{ for } |\exp(-\mat{z}_i)|>1 \rightarrow \mat{z}_i < 0,
% accuracy plot: https://www.wolframalpha.com/input/?i=log(x)+%2B+x%5E-1+-+x%5E-2%2F2+%2B+x%5E-3%2F3+-+x%5E-4%2F4+%2B+x%5E-5%2F5+-+x%5E-6%2F6+%2B+x%5E-7%2F7+-+x%5E-8%2F8+%2B+x%5E-9%2F9+-+x%5E-10%2F10+-+log(1%2Bx)+from+-1+to+10
\end{align}
and although the use of either choice would result in an ELBO lower bound, we choose the approximation based on the training label as follows;
if $y_i = 0\ \text{or}\ 1$ then we would choose the $(\mat{z}_i > 0)\ \text{or}\ (\mat{z}_i < 0)$ approximation, respectively.
We choose this because $z_i > 0$ gives a higher class conditional probability to class 0 than class 1 so this approximation would yield a tight lower bound when the training examples are correctly classified.
These approximations are plotted in \cref{fig:logistic_approx} with a first-order expansion where it is evident that the computation lower-bounds the exact computation.
Using this first-order Taylor series approximation, we can write our lower bound for the inner product between the variational distribution and the log-likelihood as follows which completes the proof,
\begin{align}
\mat{q}^T\logl &\geq
-\mat{s}^T \big(\mat{\Phi}^T \mat{y}\big) 
-\sum_{i=1}^n \left\{ \begin{array}{ll}
\mat{q}^T\exp(-\mat{z}_i)&\text{if}\ y_i = 0\\
\mat{q}^T\exp(\mat{z}_i)-\mat{q}^T\mat{z}_i&\text{if}\ y_i = 1
\end{array}\right. ,\\
&=
-\mat{s}^T \big(\mat{\Phi}^T \mat{y}\big) 
-\sum_{i=1}^n \left\{ \begin{array}{ll}
\prod_{j=1}^b \mat{q}_j^T\exp(-\phi_{ij} \widebar{\mat{w}}_j)&\text{if}\ y_i = 0\\
\prod_{j=1}^b \mat{q}_j^T\exp(\phi_{ij} \widebar{\mat{w}}_j)-\sum_{j=1}^b\mat{q}_i^T\phi_{ij} \widebar{\mat{w}}_j&\text{if}\ y_i = 1
\end{array}\right. .
\end{align}
\mbox{} \hfill $\square$

\paragraph{Remark}
\label{sec:logistic_remarks}
We expect these Taylor series approximations to admit a tight bound within and just outside of their logit domains as we can see in \cref{fig:logistic_approx}.
Equivalently, the log-likelihood approximation is accurately computed for hypotheses that correctly classify the training data when we use this lower bound, however, hypotheses that confidently misclassify training labels may be over-penalized.
This can be seen by observing how the approximations in \cref{fig:logistic_approx} significantly underestimate the exact solution when they are far outside of the approximations domain.

\paragraph{Remark} In practice, the products over $b$ terms in \Cref{thm:logistic} may result in overflow or loss of precision, however, computations can be performed in a stable manner in logit space and the LogSumExp trick~\cite{logsumexp} can be used to avoid precision loss for the sum over $n$.
\begin{comment}
To demonstrate, take truncate the $\mat{z}_i > 0$ Taylor series after $h$ terms and consider the inner product with the variational distribution (considering mean-field for simplicity)
\begin{align}
\mat{q}^T\logl + \sum_{i=1}^n y_i\mat{q}^T\mat{z}_i 
&= \sum_{k=1}^h \frac{(-1)^k}{k} \mat{q}^T\exp(-k\mat{z}_i),\\
&= \sum_{k=1}^h \frac{(-1)^k}{k} \prod_{j=1}^b \mat{q}_j^T\exp(-k\phi_{ij}\widebar{\mat{w}}_j),\\
&= \sum_{k=1}^h \frac{(-1)^k}{k} \prod_{j=1}^b \text{SumExp}\big( \log\mat{q}_j - k\phi_{ij}\widebar{\mat{w}}_j\big),\\
&= \sum_{k=1}^h \frac{(-1)^k}{k} \exp \bigg( \sum_{j=1}^b \text{LogSumExp}\big( \log\mat{q}_j - k\phi_{ij}\widebar{\mat{w}}_j\big)\bigg),\\
&= \sum_{k=1}^h (-1)^k \exp \bigg(-\log k + \sum_{j=1}^b \text{LogSumExp}\big( \log\mat{q}_j - k\phi_{ij}\widebar{\mat{w}}_j\big)\bigg),
\end{align}
which I think should be stable to compute. 
A similar scheme can be devised for the $\mat{z}_i < 0$ series expansion.
\end{comment}

\section{Proof of \Cref{thm:mixture_lb}: Mixture Distribution Entropy Lower Bound}
\label{sec:mixture_proof}
We begin by taking a Taylor series approximation of $\logq$ about $\mat{a} = \otimes_{i=1}^b \mat{a}_i$, $\mat{a}_i \in (0,1)^{\widebar{m}}$ as follows,
\begin{align}
% https://www.wolframalpha.com/input/?i=log(x)+series+expansion+about+x%3Da
\logq = \log \mat{a} + 
\sum_{k=1}^\infty \frac{(-1)^{(k+1)}}{k\mat{a}^k}\big(\mat{q}-\mat{a}\big)^k,
\end{align}
which can be represented as a sum of Kronecker product vectors once the exponents are computed explicitly. 
However, the number of terms in this sum will grow quickly with respect to the order of the Taylor series approximation.
When a first order Taylor series expansion is considered, the approximation will give a strict lower bound of $-\logq$ and consequently a lower bound of the ELBO~(\cref{eqn:ELBO2}) will be achieved.
The approximation for a linear Taylor series expansion is plotted in \cref{fig:log_approx} where it is apparent that the approximation lower-bounds the exact computation.
We consider this linear approximation for the result in \Cref{thm:mixture_lb}. 
Note that the exact computation will always be lower bounded irrespective of the location that the Taylor series is taken about, therefore, we may select the values of $\{\mat{a}_i \in (0,1)^{\widebar{m}}\}_{i=1}^b$ that maximize this lower bound, as written in the theorem statement.
We can then write our approximation of the third term from the ELBO~(\cref{eqn:ELBO2}) to complete the proof as follows
\begin{align}
-\mat{q}^T\logq &\geq
%1-\mat{q}^T\log \mat{a} -\mat{q}^T(\mat{q}/\mat{a}) \\ &=
1-\sum_{j=1}^r \alpha_j\bigg(\sum_{i=1}^b 
\mat{q}^{(j)\,T}_i \log \mat{a}_i 
+ \alpha_j\prod_{i=1}^b \mat{q}_i^{(j)\,T} \frac{\mat{q}_i^{(j)}}{\mat{a}_i}
+ 2\hspace{-2mm}\sum_{k{=}j{+}1}^r \hspace{-2mm}\alpha_k\prod_{i=1}^b \mat{q}_i^{(j)\,T} \frac{\mat{q}_i^{(k)}}{\mat{a}_i}
\bigg).
\end{align}
%\begin{align}
%\log \bigg(\sum_{j=1}^r \alpha_j^2\prod_{i=1}^b \mat{q}_i^{(j)\,T} \frac{\mat{q}_i^{(j)}}{\mat{a}_i}\bigg)
%&= \texttt{LogSumExp}\bigg\{
%2\log \alpha_j + \sum_{i=1}^b\log\bigg(\mat{q}_i^{(j)\,T} \frac{\mat{q}_i^{(j)}}{\mat{a}_i}\bigg)
%\bigg\}_{j=1}^r\\
%\log \bigg(2\sum_{j=1}^r \alpha_j\hspace{-2mm}\sum_{k{=}j{+}1}^r \hspace{-2mm}\alpha_k\prod_{i=1}^b \mat{q}_i^{(j)\,T} \frac{\mat{q}_i^{(k)}}{\mat{a}_i}
%\bigg) &= \dots
%\end{align}

\mbox{} \hfill $\square$

\paragraph{Remark}
The products over $b$ terms might seem problematic, however, we do not expect the final results to be too large to be an overflow concern.
To avoid precision loss, we compute the log of the products, which can be done stably, and then exponentiate.

\section{DIRECT Bayesian Neural Networks for Regression}
\label{sec:bnn}
In order to demonstrate DIRECT computation of the log-likelihood for a Bayesian neural network we will first perform a forward-pass through the neural network from top to bottom, however, unlike how a forward-pass is conventionally conducted in literature where the network is fixed at a specific location in the hypothesis space, we will simultaneously evaluate the neural network at \emph{all} locations in entire hypothesis space.
Consequently, a forward-pass through the neural network with our $n$-point training set will give us $\widebar{m}^b \times n$ values.

\paragraph{Nomenclature and Neuron Structure}
At all points in the forward-pass we can represent the internal (or final) state of the neural network with a special structure which is a sum of Kronecker product vectors as follows for $i=1,\dots,\text{(number of neurons in the layer)}$, and $l=1,\dots,n$,
\begin{align} \label{eqn:nn_state}
\mat{u}^{(i)}_l = \sum_{j=1}^h c_{jl} \bigotimes_{k=1}^b \mat{g}^{(i)}_{jk},
\end{align}
where 
$\mat{U}^{(i)} = \{\mat{u}_l^{(i)}\}_{l=1}^n \inR{\widebar{m}^b \times n}$, $\mat{u}_l^{(i)} \inR{\widebar{m}^b}$ denotes the internal state of the $i$th neuron of the current layer, and 
both $\mat{G}^{(i)} = \{\{ \mat{g}^{(i)}_{jk}\}_{j=1}^h \}_{k=1}^b \inR{h\times b \times \widebar{m}}$, $\mat{g}^{(i)}_{jk} \inR{\widebar{m}}$ and
$\mat{C} \inR{h \times n}$ change as we move from one layer to the next.
$h$ depends on the network architecture and it is constant throughout a layer but grows as we observe deeper layers.
Using this nomenclature it is evident that we can compactly represent the internal state of any location within the neural network while we compute our forward pass.

The following image denotes the structure of a neuron that we will use in our neural network.
\begin{center}
\begin{tikzpicture}[
init/.style={
	draw,
	circle,
	inner sep=2pt,
	font=\Huge,
	join = by -latex
},
squa/.style={
	draw,
	inner sep=2pt,
	font=\Large,
	join = by -latex
},
start chain=2,node distance=13mm
]
\node[on chain=2] 
(x2) {};
\node[on chain=2,join=by o-latex] 
{$w_2$};
\node[on chain=2,init] (sigma) 
{$\displaystyle\Sigma$};
\node[on chain=2,init,label=above:{\parbox{2cm}{\centering Activation \\ function}}]   
{$\sim$};
\node[on chain=2,join=by -latex] 
{Layer Output};
\begin{scope}[start chain=1]
\node[on chain=1] at (0,1.5cm) 
(x1) {};
\node[on chain=1,join=by o-latex] 
(w1) {$w_1$};
\end{scope}
\begin{scope}[start chain=3]
\node[on chain=3] at (0,-1.5cm) 
(x3) {};
\node[on chain=3,label=below:Latent Variables,join=by o-latex] 
(w3) {$w_3$};
\end{scope}
\node[label=above:\parbox{2cm}{\centering Bias}] at (sigma|-w1) (b) {};

\draw[-latex] (w1) -- (sigma);
\draw[-latex] (w3) -- (sigma);
\draw[o-latex] (b) -- (sigma);

\draw[decorate,decoration={brace,mirror}] (x1.north west) -- node[left=10pt] {\parbox{0.8cm}{Layer\\ Inputs}} (x3.south west);
\end{tikzpicture}
\end{center}
For clarity of illustration, we will not discuss the bias term although this can be easily added by associating a latent variable with a layer input that is fixed to unity.
In our discussion, we will break the computation of the neuron into two stages;
the first will involve multiplication of the layer inputs with the latent variables as well as the summation, and
the second stage will involve passing this summation through a non-linear activation function.

\paragraph{Multiplication with Latent Variables and Summation}
Our computational neurons begin by multiplying the layer inputs with a specific latent variable and then summing up these values. 
Assuming we are conducting a forward-pass moving deeper into the network and are currently at the "layer inputs" location in our computational neuron figure, the internal state for the $i$th input is denoted by $\mat{U}^{(i)} \inR{\widebar{m}^b \times n}$ whose structure is defined in \cref{eqn:nn_state}.
We must multiply this state by all possible values of the corresponding latent variable, which we will assume is indexed as the $p$th of our $b$ latent variables.
We can easily perform this multiplication as follows for $l=1,\dots,n$
\begin{align}
\mat{u}'^{(i)}_l 
&= \bigg(\sum_{j=1}^h c_{jl} \bigotimes_{k=1}^b \mat{g}^{(i)}_{jk} \bigg)
\odot \mat{W}[p,:]^T,\\
&= \sum_{j=1}^h c_{jl} \bigg(\bigotimes_{k=1}^{p-1} \mat{g}^{(i)}_{jk} \bigg) 
\otimes \big(\mat{g}^{(i)}_{jp} \odot \widebar{\mat{w}}_p\big)
\otimes \bigg(\bigotimes_{k=p+1}^{b} \mat{g}^{(i)}_{jk} \bigg)
=\sum_{j=1}^h c_{jl} \bigotimes_{k=1}^b \mat{g}'^{(i)}_{jk},
\label{eqn:mult_var}
\end{align}
where $\odot$ denotes element-wise multiplication, and
we have taken advantage of the Kronecker product structure of the rows of $\mat{W}$ as depicted in \cref{eqn:w_grid}.
Finally, the summing operation is straightforward for our computational neuron. 
It simply involves summing the multiplied inputs from each layer input as follows,
\begin{align}
\sum_{i=1}^\text{num. inputs} \sum_{j=1}^h c_{jl} \bigotimes_{k=1}^b \mat{g}'^{(i)}_{jk}.
\label{eqn:neuron_sum}
%=\sum_{j=1}^h c_{jl} \sum_{i=1}^\text{num. inputs} \bigotimes_{k=1}^b \mat{g}'^{(i)}_{jk}.
\end{align}
At this point we would update $h$, $\mat{G}$ and $\mat{C}$ to convert this double summation into a single summation before passing through the non-linear activation function, as we will discuss next.

\paragraph{Quadratic Activation}
We will use a quadratic activation function for our neural network.
Any other non-linear activation could be used, however, we choose the quadratic since it allows a more compact representation of internal state of the network to be maintained, i.e. allows for a small $h$ versus other non-linear activations.
Again assuming that the current state at the $i$th neuron is defined by $\mat{U}^{(i)}$,
the output for the activation function for the $i$th neuron is as follows for $l=1,\dots,n$
\begin{align}
\mat{u}'^{(i)}_l 
&= \mat{u}^{(i)}_l \odot \mat{u}^{(i)}_l =
\bigg(\sum_{j=1}^h c_{jl} \bigotimes_{k=1}^b \mat{g}^{(i)}_{jk} \bigg) 
\odot
\bigg(\sum_{j=1}^h c_{jl} \bigotimes_{k=1}^b \mat{g}^{(i)}_{jk} \bigg), \\
&= \sum_{j=1}^h c_{jl}^2 \bigotimes_{k=1}^b \mat{g}^{(i)}_{jk} \odot \mat{g}^{(i)}_{jk}
+ 2 \sum_{j=1}^h \sum_{p=1}^{j-1} c_{jl}c_{pl} \mat{g}^{(i)}_{jk} \odot \mat{g}^{(i)}_{pk},
\label{eqn:activation}
\end{align}
and at this point we would update $h$, $\mat{G}$ and $\mat{C}$ to convert this double summation into a single summation to represent the internal state compactly before moving deeper.

\paragraph{Forward-Pass Algorithm}
Using the previously defined operations, we can summarize the forward-pass procedure in \cref{alg:forward_pass}.
Note that \cref{alg:forward_pass} is simplified for clarity of presentation.
The computations involved could be performed far more efficiently and in a more stable manner.
For example, the vast majority of entries in the $\mat{G}$ matrices are unity, so identifying this could massively decrease storage and computational requirements.
Additionally, $\widetilde{\mat{C}}$ evidently has a Kronecker product structure which could be carefully exploited to yield benefits for very wide neural networks.
For stability, all matrices could be represented by storing both the sign and logarithm of all entries. 
For deep networks, this could be advantageous to avoid precision loss.
Nonetheless, we will proceed with the algorithm as presented, for purposes of clarity.

\begin{varalgorithm}{\texttt{forward\_pass}}
	\caption{
	Perform a forward pass for through the neural network using the entire training set and simultaneously computing the outputs for all $m=\widebar{m}^b$ points in the hypothesis space.
	\texttt{mult\_var} multiplies the current state with the appropriate latent variable as is done in \cref{eqn:mult_var},
	\texttt{neuron\_sum} computes the neuron sum as is done in \cref{eqn:neuron_sum}, and
	\texttt{activation} computes the non-linear activation function as is done in \cref{eqn:activation}.
	All the pseudo-functions defined take $\mat{G}$ and/or $\mat{C}$ and perform the necessary computations with those inputs.
	We omit latent-variable indexing values for clarity of presentation.}
	\label{alg:forward_pass}
	\begin{algorithmic}
		\STATE {\bfseries Input:} $\mat{X} \inR{n\times d}$
		\STATE {\bfseries Output:}  $\mat{C} \inR{h\times n}$ \& $\mat{G} \inR{h\times b \times \widebar{m}}$ which define state $\mat{U} \inR{\widebar{m}^b \times n}$ in \cref{eqn:nn_state}
		\STATE $\mat{C} = \mat{X}^T$, $\quad$ 
		$\mat{G}^{(i)} = \texttt{ones}(1 \times b \times \widebar{m}),\ i=1,\dots,d$
		\FOR{each layer} 
		\STATE $\widetilde{\mat{C}} = \texttt{neuron\_sum}(\{\mat{C}\}_1^\text{num. inputs})= \mat{1}_\text{num. inputs} \otimes \mat{C}$
		\FOR{$j=1$ {\bfseries to} num. neurons in layer}
		\FOR{$i=1$ {\bfseries to} num. inputs to layer}
		\STATE $\mat{G}'^{(i)} = \texttt{mult\_var}(\mat{G}^{(i)})$ \hfill multiplication with the appropriate row of $\mat{W}$
		\ENDFOR
		\STATE $\widetilde{\mat{G}}^{(j)} = \texttt{neuron\_sum}(\mat{G}'^{(1)},\dots,\mat{G}'^{(\text{num. inputs})})$ \hfill
		sum operation for the current ($j$th) neuron
		\IF{{\bfseries not} last layer} %{no activation function on last layer}
		\STATE $\widetilde{\mat{G}}^{(j)},\ \widetilde{\mat{C}} = \texttt{activation}(\widetilde{\mat{G}}^{(j)}, \widetilde{\mat{C}})$
		\ENDIF
		\ENDFOR
		\STATE ${\mat{C}} = \widetilde{\mat{C}}$, $\quad {\mat{G}}^{(j)} = \widetilde{\mat{G}}^{(j)}, j=1,\dots,\text{num. neurons in layer}$ \hfill update variables
		\ENDFOR
		\STATE $\mat{G} = \mat{G}^{(1)}$\hfill only one neuron in the last (output) layer, so remove indexing
	\end{algorithmic}
\end{varalgorithm}

\paragraph{ELBO Computation}
Computation of the ELBO will proceed similarly to the GLM regression model in \cref{sec:glm}, however, there are several differences since we no longer have constant basis functions so our state representation is more complicated.
We will again assume a Gaussian noise model for the observed responses and will again place a prior over the Gaussian variance.
We can then modify \cref{eqn:glm_qTlogl} which focuses on the ELBO term related to the log-likelihood as follows
\begin{align} \label{eqn:nn_qTlogl}
(\mat{q}_{\sigma} \otimes \mat{q})^T \logl =
-\frac{n}{2}\mat{q}_{\sigma}^T \log\mat{\sigma}^2
-\frac{1}{2}\big(\mat{q}_{\sigma}^T \mat{\sigma}^{-2}\big) \big(\mat{q}^T \{(\mat{y} - \mat{U}[i,:]^T)^T (\mat{y} - \mat{U}[i,:]^T)\}_{i=1}^m \big),
\end{align}
where we assume that we have already conducted \cref{alg:forward_pass} such that the state $\mat{U}$ represents the output of the neural network.
We will now focus on computing the inner product involving the variational distribution over the $\mat{w}$ variables, $\mat{q}$, which we can break into three terms as follows,
\begin{multline}
\mat{q}^T\{(\mat{y} - \mat{U}[i,:]^T)^T (\mat{y} - \mat{U}[i,:]^T)\}_{i=1}^m
=\\ \mat{y}^T\mat{y} 
- 2\mat{q}^T\{\mat{y}^T \mat{U}[i,:]^T\}_{i=1}^m 
+ \mat{q}^T\{\mat{U}[i,:]\mat{U}[i,:]^T\}_{i=1}^m, \label{eqn:nn_terms}
%&= \sum_{i=1}^n y_i^2 - \sum_{i=1}^n 2 y_i \mat{\Phi}[i,:] \mat{w} + \sum_{i=1}^n(\mat{\Phi}[i,:] \mat{w})^2 
\end{multline}
for which the first term is trivial to compute as written since it does not depend on the latent variables.
We now demonstrate how the second and third terms can be computed, recalling we assume $\mat{q}$ is a mean-field variational distribution (although we can extend beyond mean-field using the techniques discussed in \cref{sec:extensions}).
Considering the second term in \cref{eqn:nn_terms}, we can write
\begin{align} \nonumber
\mat{q}^T\{\mat{y}^T \mat{U}[i,:]^T\}_{i=1}^m
&= 
\mat{q}^T \bigg( \sum_{k=1}^{n} y_k \sum_{j=1}^h c_{jk} \bigotimes_{i=1}^b \mat{g}_{ij}\bigg)
= \sum_{j=1}^h \bigg(\sum_{k=1}^{n} y_k c_{jk}\bigg) \prod_{i=1}^b \mat{q}_i^T\mat{g}_{ij},\\
&= \sum_{j=1}^h p_j \prod_{i=1}^b \mat{q}_i^T\mat{g}_{ij}, \label{eqn:nn_term2}
\end{align}
where we have used the short-hand notation $\mat{p} = \{\sum_{k=1}^{n} y_k c_{jk}\}_j \inR{h}$.
Finally, considering the third term in \cref{eqn:nn_terms}, we can write
\begin{align}
\mat{q}^T\{\mat{U}[i,:]\mat{U}[i,:]^T\}_{i=1}^m
&= \mat{q}^T \sum_{i=1}^n  \mat{u}_i \odot \mat{u}_i
= \mat{q}^T\sum_{i=1}^n \sum_{j=1}^h \sum_{k=1}^h c_{ji}c_{ki} \bigotimes_{l=1}^b \big(\mat{g}_{jl} \odot \mat{g}_{kl}\big),\\
&= \sum_{j=1}^h \sum_{k=1}^h \bigg(\sum_{i=1}^n c_{ji}c_{ki}\bigg) \prod_{l=1}^b \mat{q}^T_l \big(\mat{g}_{jl} \odot \mat{g}_{kl}\big),\\
&= \sum_{j=1}^h \sum_{k=1}^h v_{jk} \prod_{l=1}^b \mat{q}^T_l \big(\mat{g}_{jl} \odot \mat{g}_{kl}\big), \label{eqn:nn_term3}
\end{align}
where we define $\mat{V} = \{\sum_{i=1}^n c_{ji}c_{ki}\}_{j,k} \inR{h\times h}$.
Substituting \cref{eqn:nn_term2} and \cref{eqn:nn_term3} into \cref{eqn:nn_terms}, we can now compute the inner product between the variational distribution and the log-likelihood in \cref{eqn:nn_qTlogl}.
The other terms required to compute the ELBO can be seen in \cref{eqn:ELBO_w_sigma}, and the computation of these other terms do not differ from the case of the generalized linear regression model.
So we can now tractably compute the ELBO for our DIRECT Bayesian neural network.

We can pre-compute the terms $\mat{y}^T\mat{y}$, $\mat{p}$, and $\mat{V}$ before training begins (since these do not depend on the variational parameters)
such that the final complexity of the DIRECT method is \emph{independent} of the number of training points, making the proposed techniques ideal for massive datasets.
Also, it is evident that each of these pre-computed terms can easily be updated as more data is observed making the techniques amenable to online learning applications.
If we assume a neural network with $\ell$ hidden layers and an equal distribution of latent variables between layers, the computational complexity of the ELBO computations are
\order{\ell \widebar{m}(b/\ell)^{4\ell}}.
This can be seen by observing \cref{eqn:nn_term3} and noting that $h = $\order{(b/\ell)^{2\ell}}, and that only \order{\ell} of the vectors in
$\{\mat{g}_{jl}\}_{l=1}^b$ are not unity for any value of $j=1,\dots,h$, allowing computations to be saved.

\begin{comment}
Unfortunately, it is not so clear how a Bayesian deep neural network can be used for logistic regression since $\exp(-\mat{z}_i)$ would not have a compact form as is exploited for the logistic regression GLM in \cref{sec:logistic}.
In this case, we clarify that $\mat{z}_i\inR{m}$ are the logits at the final layer of the neural network for the $i$th train point.
\end{comment}

\FloatBarrier
\section{UCI Regression Studies Setup \& Additional Results}
\label{sec:uci_setup}
Considering regression datasets from the UCI repository, we report the mean and standard deviation of the root mean squared error (RMSE) from 10-fold cross validation%
\footnote{90\% train, 10\% test per fold. We use folds from \url{https://people.orie.cornell.edu/andrew/code}}.
Also presented is the mean training time per fold on a machine with two E5-2680 v3 processors and 128Gb of RAM, and the expected sparsity (percentage of zeros) within a posterior sample.
Using a generalized linear model, we consider $b=2000$ random Fourier features of a squared-exponential kernel with automatic relevance determination~\cite{rahimi_rff}.
Before generating the features, we initialize the kernel hyperparameters including the prior variance $\sigma_w^2$ and the Gaussian noise variance $\sigma^2$ by maximizing the marginal likelihood of an exact Gaussian process constructed on $\min(n,1000)$ points randomly selected from the dataset~\cite{rasmussen_gpml}.
All discretely relaxed models~(containing ``DIRECT''), only have support at 
$w \in \texttt{linspace}(-3\sigma_w,3\sigma_w,\widebar{m}{=}15)$, allowing $\mat{w}$ to be stored as 4-bit quantized integers.
%We fix $\sigma^2$ to be a single value since a prior over this value cannot be computed analytically for our baseline (although \cref{sec:glm} discussed how we can easily place a prior distribution over $\sigma^2$ for our discrete model in practice).

For REPARAM we perform doubly stochastic optimization using a mini-batch size of 100 and using 10 Monte Carlo samples for the gradient estimates at each iteration. 
For datasets with $n<3000$ we optimize for 1000 iterations and we optimize for 10000 iterations for all larger datasets.
This model was implemented in Edward~\cite{edward}.
For the DIRECT mean-field model we use an L-BFGS optimizer~\cite{nocedal_lbfgs} and run until convergence, or 1000 iterations are reached.
For the DIRECT 5-mixture model we perform stochastic gradient descent using $t=3000$ Monte Carlo samples for the entropy gradient estimator \cref{eqn:entropy_surrogate}.

For the DIRECT mean-field model we initialize the variational distribution to the prior.
For the DIRECT mixture models,
we first run the mean-field model and then initialize each mixture component to be randomly perturbed from the mean-field solution, and we initialize $\mat{a}$ to the mean-field solution.
We initialize the mixture probabilities to be constant.

For predictive posterior mean computations, we use the exact computation presented in \cref{eqn:post_mean} for both the DIRECT  and mixture models.
For REPARAM, we approximate the mean by sampling the variational distribution using 1000 samples.

In \cref{tbl:uci_biased_results} we consider again an unfactorized mixture variational distribution, however, we maximize the ELBO lower bound derived in \Cref{thm:mixture_lb}.
Since the ELBO gradients are deterministic, we again use an L-BFGS optimizer for training.
In addition to the $150,000$ variational parameters used by the DIRECT 5-Mixture SGD model in \cref{tbl:uci_results}, computing the ELBO lower bound involves the simultaneous optimization of $\mat{a}$, adding $30,000$ additional optimization parameters.

\begin{table}
	\centering
	\setlength\tabcolsep{1.5pt} % default value: 6pt
	\scriptsize%\hspace{-1cm}%
	\begin{tabular}{@{}|lll|lll|@{}}\toprule%&&&&&&
&&&\multicolumn{3}{c|}{Discrete 4-bit Prior}\\%&\\%&&
&&&\multicolumn{3}{c|}{DIRECT 5-Mixture ELBO-LB}\\%&\\%&&
Dataset&$n$&$d$&Time&RMSE&Sparsity\\%&\\%
\midrule%&&&&&&
challenger&23&4&15&$0.528 \pm 0.243$&16\%\\%&\\%
fertility&100&9&15&$0.16 \pm 0.04$&16\%\\%&\\%
%slump&103&7&5&$53.77 \pm 11.437$&100\%\\%&\\%
automobile&159&25&24&$0.137 \pm 0.053$&47\%\\%&\\%
servo&167&4&24&$0.282 \pm 0.067$&32\%\\%&\\%
cancer&194&33&17&$23.344 \pm 3.414$&18\%\\%&\\%
hardware&209&7&24&$0.492 \pm 0.117$&46\%\\%&\\%
yacht&308&6&5&$0.23 \pm 0.077$&96\%\\%&\\%
autompg&392&7&24&$2.624 \pm 0.339$&29\%\\%&\\%
housing&506&13&24&$2.782 \pm 0.324$&37\%\\%&\\%
forest&517&12&15&$1.361 \pm 0.159$&16\%\\%&\\%
stock&536&11&233&$0.011 \pm 0.002$&98\%\\%&\\%
pendulum&630&9&6&$1.36 \pm 0.227$&68\%\\%&\\%
energy&768&8&5&$3.116 \pm 0.218$&99\%\\%&\\%
concrete&1030&8&19&$5.571 \pm 0.665$&81\%\\%&\\%
solar&1066&10&24&$0.799 \pm 0.192$&22\%\\%&\\%
airfoil&1503&5&16&$2.175 \pm 0.32$&46\%\\%&\\%
wine&1599&11&24&$0.486 \pm 0.047$&50\%\\%&\\%
gas&2565&128&5&$0.204 \pm 0.053$&84\%\\%&\\%
skillcraft&3338&19&78&$0.253 \pm 0.017$&97\%\\%&\\%
sml&4137&26&7&$0.675 \pm 0.044$&57\%\\%&\\%
parkinsons&5875&20&8&$0.642 \pm 0.06$&13\%\\%&\\%
%pumadyn&8192&32&5&$0.997 \pm 0.032$&100\%\\%&\\%
poletele&15000&26&24&$13.728 \pm 0.447$&16\%\\%&\\%
elevators&16599&18&5&$0.124 \pm 0.003$&99\%\\%&\\%
%kin40k&40000&8&8&$0.413 \pm 0.018$&69\%\\%&\\%
protein&45730&9&16&$0.62 \pm 0.007$&76\%\\%&\\%
kegg&48827&20&6&$0.222 \pm 0.01$&95\%\\%&\\%
ctslice&53500&385&6&$6.036 \pm 0.163$&19\%\\%&\\%
keggu&63608&27&6&$0.139 \pm 0.004$&87\%\\%&\\%
3droad&434874&3&14&$10.487 \pm 0.075$&40\%\\%&\\%
song&515345&90&8&$0.502 \pm 0.002$&31\%\\%&\\%
buzz&583250&77&9&$1.009 \pm 0.004$&82\%\\%&\\%
electric&2049280&11&5&$0.593 \pm 0.036$&99.6\%\\%&\\%
\bottomrule \end{tabular}%&&&&&&

	\caption{Using a mixture variational distribution along with the the ELBO lower bound presented in \Cref{thm:mixture_lb}, we present the mean and standard deviation of test error, average training time, and average expected sparsity of a posterior sample from 10-fold cross validation on UCI regression datasets.}
	\label{tbl:uci_biased_results}
\end{table}

\begin{comment}
\section{Extra Stuff}
\begin{proposition}
	The element-wise product of two Kronecker product vectors 
	$\mat{k} = \otimes_{i=1}^b \mat{k}^{(i)}$, and
	$\mat{a} = \otimes_{i=1}^b \mat{a}^{(i)}$
	can be written as a Kronecker product vector;
	\begin{align}
	\mat{a} \odot \mat{k} = \bigotimes_{i=1}^b \mat{a}^{(i)} \odot \mat{k}^{(i)}
	\end{align}
	where
	$\mat{a}^{(i)} \inR{\widebar{m}}$, $\mat{a} \inR{\widebar{m}^b}$, 
	$\mat{k}^{(i)} \inR{\widebar{m}}$, and $\mat{k} \inR{\widebar{m}^b}$.
\end{proposition}

\paragraph{Mixture probability distribution} with $r$ components can be written
\begin{align}
% this was inspired by this paper eqn 2 https://www.ncbi.nlm.nih.gov/pmc/articles/PMC3630378/
\Pr(w_1{=}\widebar{w}_1[c_1], \dots, w_b{=}\widebar{w}_b[c_b]) = \sum_{i=1}^r \alpha_i \prod_{j=1}^b q^{(i)}_{jc_j},
\end{align}
\end{comment}
\end{document}